\documentclass{article}

 \usepackage[eandd,final]{neurips_2026}

\usepackage[utf8]{inputenc} % allow utf-8 input
\usepackage[T1]{fontenc}    % use 8-bit T1 fonts
\usepackage{hyperref}       % hyperlinks
\usepackage{url}            % simple URL typesetting
\usepackage{booktabs}       % professional-quality tables
\usepackage{amsfonts}       % blackboard math symbols
\usepackage{nicefrac}       % compact symbols for 1/2, etc.
\usepackage{microtype}      % microtypography
\usepackage{xcolor}         % colors
\usepackage{amsmath}
\usepackage{graphicx}
\usepackage{caption}       % \captionof{figure}{...} inside minipages
\usepackage{algorithm}
\usepackage{algorithmic}
\usepackage{wrapfig}       % in-text figure wrapping for the intro teaser
\usepackage[normalem]{ulem} % \uline{...} for Q&A bookend formatting (Rule 18)

\newcommand{\method}{GRADE}
\newcommand{\methodOurs}{\textbf{\method{} (ours)}}

\newcommand{\backboneA}{Llama-3.1-8B}
\newcommand{\backboneB}{Qwen3-8B}
\newcommand{\backboneC}{Gemma-2-9B}
\newcommand{\backboneAll}{\backboneA, \backboneB, and \backboneC}
\newcommand{\backboneLegacy}{LLaMA-2-7B}

\newcommand{\WorstDelta}{Worst-$\Delta$}
\newcommand{\MeanDelta}{Mean-$\Delta$}

\newcommand{\rstar}{\ensuremath{r^{\star}}}
\newcommand{\cosintra}{\ensuremath{\cos_{\mathrm{intra}}}}
\newcommand{\cosinter}{\ensuremath{\cos_{\mathrm{inter}}}}
\newcommand{\Lprobe}{\ensuremath{\overline{L}_{\mathrm{probe}}}}
\newcommand{\Lprobet}[1]{\ensuremath{\overline{L}_{\mathrm{probe}}^{(#1)}}}
\newcommand{\epsrel}{\ensuremath{\varepsilon_{\mathrm{rel}}}}
\newcommand{\Glora}{\ensuremath{G^{(t)}_{\mathrm{LoRA}}}}
\newcommand{\GFCintra}{\ensuremath{\mathrm{GFC}_{\mathrm{intra}}}}
\newcommand{\GFCinter}{\ensuremath{\mathrm{GFC}_{\mathrm{inter}}}}

\title{GRADE: A Data-Centric PEFT Recipe for Small Language Models with State-Coupled Gradient-Aligned Selection and a Step-Level Admission Gate}
\title{Data-centric Small Language Model Learning: \\A Gradient Admission Perspective}
\title{Which and When to Admit: Gradient Admission for Data-Centric Small Language Model Finetuning}
\author{%
Hongyu Cao\textsuperscript{1},
Yanchi Liu\textsuperscript{1},
Kunpeng Liu\textsuperscript{3},
Xujiang Zhao\textsuperscript{4},
\\
Wei Cheng\textsuperscript{2},
Zhengzhang Chen\textsuperscript{2},
Yanjie Fu\textsuperscript{1},
Haifeng Chen\textsuperscript{2}
\\
\textsuperscript{1}Arizona State University
\\
\textsuperscript{2}NEC Laboratories America
\\
\textsuperscript{3}Clemson University,
\textsuperscript{4}Meta AI
}
\begin{document}
% Review Token: FfYNHsgix7w984G9leZ5pDf513gdjcFSzgBLLxHScHU

\maketitle

\begin{abstract}
LoRA fine-tuning adapts small language models (SLMs) to heterogeneous instruction data by confining updates to a low-rank parameter subspace. We show that this paradigm is structurally fragile along three axes:
(i) \textbf{gradient conflict}, where heterogeneous supervision produces incompatible update directions that partially cancel within the constrained LoRA subspace;
(ii) \textbf{state mismatch}, where static or one-shot data selection fails to track the evolving multi-task learning direction during training; and
(iii) \textbf{subspace saturation}, where later optimization steps overwrite previously useful directions once the low-rank update space is exhausted.
We argue that effective LoRA adaptation requires deciding \emph{which} data-induced gradients are admitted into the subspace and \emph{when}, rather than unconditionally accumulating every update. To this end, we propose \textbf{GRADE} (\textbf{GR}adient-\textbf{A}ligned \textbf{D}ata-centric r\textbf{E}cipe), a data-centric LoRA fine-tuning framework with two coupled components.
First, a \emph{state-aware gradient-aligned selector} continuously re-scores candidate samples by their directional agreement with the evolving multi-task gradient field, admitting only samples whose updates are compatible with the current model state.
Second, a \emph{self-calibrating step-level admission gate} commits optimizer updates only when doing so does not induce destructive overwrite near subspace saturation.
Across three current-generation backbones and a heterogeneous seven-dataset instruction pool, GRADE improves both accuracy and robustness over strong data-selection and PEFT-stabilization baselines. Notably, GRADE is the only method that achieves strictly positive gains over standard LoRA on every evaluated architecture, while also producing more coherent gradient trajectories and reducing destructive overwrite during training.
Together, these results indicate that SLM fine-tuning depends not only on \emph{what} data are selected, but on \emph{which} gradients are allowed to enter and persist in the constrained adaptation subspace.
\end{abstract}

\vspace{-0.45cm}
\section{Introduction}
\label{sec:intro}
\vspace{-0.2cm}

Small language models (SLMs) are commonly adapted by fine-tuning lightweight LoRA adapters on top of frozen backbones, using heterogeneous instruction data collected from reasoning, question answering, summarization, dialogue, and code-like tasks~\citep{pmlr-v139-killamsetty21a,pmlr-v235-xia24c,wang-etal-2025-clusterucb,hu2022lora}. 
This data-centric LoRA finetuning paradigm is attractive because it is modular, inexpensive, and easy to deploy across backbones. 
However, its empirical behavior is often unstable: adding more data does not always improve performance, a mixture that helps one backbone may hurt another, and continued fine-tuning can overwrite rather than consolidate earlier gains~\citep{yu2020pcgrad,liu2021cagrad}. 
This raises a basic question: \textit{which data-induced gradients should be allowed to enter a constrained LoRA subspace, and when should their updates be committed?}

\begin{wrapfigure}[16]{r}{0.35\linewidth}
  \vspace{-0.6em}
  \centering
  \includegraphics[width=\linewidth]{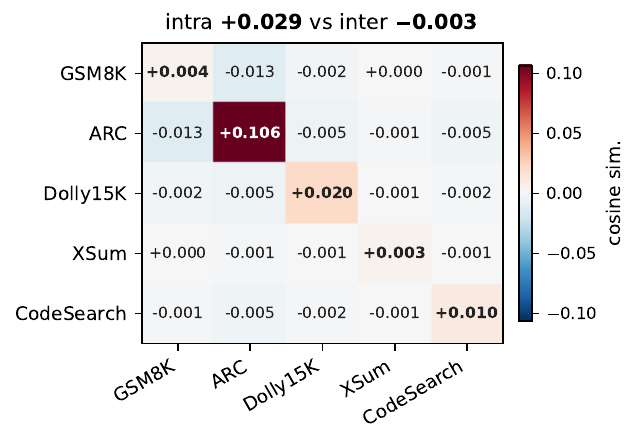}
  \caption{\small Per-sample LoRA gradient cosine on Gemma-2-9B: training-pool datasets have inconsistent gradient directions (intra-dataset $+0.029$, inter-dataset $-0.003$). Setup in App.~\ref{app:hyperparameters}.}
  \label{fig:gradient_conflict}
  \vspace{-0.5em}
\end{wrapfigure}

We observe that the core difficulty is not simply low data quality or class imbalance, but incompatible update directions induced by heterogeneous supervision. 
Figure~\ref{fig:gradient_conflict} shows that, on Gemma-2-9B, per-sample LoRA gradients are only weakly aligned within datasets and become nearly orthogonal 
or conflicting across datasets. 
Under full-parameter fine-tuning, such conflicts may be partially absorbed across many degrees of freedom. 
Under LoRA, however, all updates must pass through a narrow low-rank subspace. 
As a result, incompatible gradients partially cancel each other before they accumulate, and once the subspace approaches saturation~\citep{aghajanyan2021intrinsic,zhang2023adalora}, later steps can overwrite directions that were previously useful. 
We identify three recurring failure modes. 
First, \textit{gradient conflict}: heterogeneous data induce update directions that cannot be simultaneously represented in the restricted LoRA subspace. 
Second, \textit{state mismatch}: one-shot data selection scores at the beginning of model training become stale as the model state and the useful multi-task learning direction evolve. 
Third, \textit{subspace overwrite}: after useful low-rank directions have been occupied, additional updates may damage earlier gains rather than expand capability.

\textbf{Our Perspective.}
We argue that these failure modes share a common structural cause: existing finetuning pipelines treat training data as examples to be selected, but not as gradients to be admitted. 
Standard LoRA implicitly admits every selected sample and commits every resulting update. 
Data-selection methods improve the quality of the examples that enter training, but usually score them against a fixed reference and do not track whether their gradients remain compatible with the evolving model state~\citep{pmlr-v139-killamsetty21a,pmlr-v235-xia24c,wang-etal-2025-clusterucb}. 
Stability-aware LoRA finetuning methods~\citep{chang-etal-2025-lora,wang2025ctrlora} regularize updates after they are formed, without questioning whether the data that produced them should have been admitted. 
Adaptive LoRA finetuning methods~\citep{zhang2023adalora,liu-etal-2024-alora,zhang-etal-2025-sensitivity,buehler2024xlora,feng2025moa,kunwar2025ttloramoe} are model-centric: change low-rank subspace to adjust learning capacity, but assume training samples are all reliable. 
%Stability-aware LoRA methods regularize or reshape updates after a batch has already been formed~\citep{chang-etal-2025-lora,wang2025ctrlora}. 
%Adaptive LoRA methods change the low-rank subspace or allocate capacity differently~\citep{zhang2023adalora,liu-etal-2024-alora,zhang-etal-2025-sensitivity,buehler2024xlora,feng2025moa,kunwar2025ttloramoe}. 
These approaches differ in mechanism, but they share the same assumption: once data are selected, their gradients are admissible. 
Under heterogeneous supervision and restricted update capacity, this assumption breaks down. 
The fix requires two coupled decisions: (i) admit only samples whose gradients align with the current aggregate multi-task direction; and (ii) commit an update only when it remains constructive relative to the model's recent trajectory.

\textbf{Proposed Framework.}
We introduce \method{}, a gradient-admission framework for data-centric LoRA fine-tuning. 
First, \method{} performs \textit{state-aware gradient-aligned data admission}. 
At each iteration, candidate samples are re-scored according to their directional agreement with an evolving multi-task reference direction, so the selected data track what the current model still needs rather than what appeared useful at initialization. 
Second, \method{} applies a \textit{self-calibrating step-level admission gate}. 
Even after aligned samples are selected, the resulting base-LoRA update is committed only when its probe loss remains below a running exponential moving average baseline. 
When the update no longer improves relative to the model's own recent trajectory, the optimizer step is skipped. 
Together, these two components turn data-centric finetuning from unconditional accumulation into controlled gradient admission under subspace constraints.

\textbf{Contributions.} 
i) We identify three structural fragilities in data-centric LoRA fine-tuning of SLMs: gradient conflict, state mismatch, and subspace overwrite, and show that they arise from uncontrolled gradient admission under restricted update capacity.
ii) We propose \method{}, a gradient-admission framework that couples state-aware gradient-aligned sample admission with a self-calibrating step-level update gate.
iii) We evaluate \method{} on three current-generation backbones (Llama-3.1-8B, Qwen3-8B, and Gemma-2-9B) under a uniform heterogeneous supervision training setup. 
The key result is not only improved average accuracy, but robustness across architectures: \method{} is the only method that achieves both the highest cross-architecture mean gain and a strictly positive worst-case gain over LoRA. 
\method{} can reduce gradient conflicts across datasets and prevent destructive overwrite as the low-rank subspace fills up.

\section{Related Work}
\label{sec:related_work}

\textbf{Data-centric fine-tuning of small language models.}
A growing line of work treats data composition, which examples to admit and in what proportion, as the primary lever for adapting small language models to a target capability~\citep{liu2026continuousoptimizationfeatureselection}.
Gradient-aligned data selection methods such as GRAD-MATCH select subsets whose aggregate gradients approximate the full or validation gradient~\citep{pmlr-v139-killamsetty21a}; LESS builds a reusable low-dimensional gradient datastore and selects examples by influence-aware similarity to a small target set~\citep{pmlr-v235-xia24c}; ClusterUCB reduces scoring cost via bandit-style allocation over gradient clusters~\citep{wang-etal-2025-clusterucb}.
Alternative compatibility proxies include difficulty and forgetting scores~\citep{toneva2019example,paul2021deep} and diversity-aware gradient selection~\citep{pan2024gdig}, which target hardness, instability, or coverage rather than directional consistency~\citep{cao2026rethinkingdataaugmentationcovariate}.
\vspace{-0.15cm}

\textbf{LoRA finetuning under restricted and adaptive subspaces.}
LoRA freezes the backbone and parameterizes updates as low-rank matrices~\citep{hu2022lora}.
Several methods allocate the low-rank budget adaptively (AdaLoRA~\citep{zhang2023adalora}, ALoRA~\citep{liu-etal-2024-alora}, Sensitivity-LoRA~\citep{zhang-etal-2025-sensitivity}) or improve the parameterization itself (DoRA~\citep{liu2024dora}, LoRA+~\citep{hayou2024loraplus}, PiSSA~\citep{meng2024pissa}, GaLore~\citep{zhao2024galore}); a stability-aware line regularizes the update trajectory after a batch is formed (LoRA-MGPO~\citep{chang-etal-2025-lora}, CtrLoRA~\citep{wang2025ctrlora}).
These methods reshape the optimization geometry within a fixed subspace and assume the observed training signal faithfully reflects the underlying capacity requirement; under heterogeneous supervision that assumption fails because the signal is shaped by conflicting per-sample gradients before reaching the optimizer.
\method{} is orthogonal in that it asks whether the candidate data should be admitted in the first place, and once admission is filtered, whether each individual step continues to be constructive and remains compatible with any of these parameterizations as a drop-in replacement for vanilla LoRA.
\vspace{-0.15cm}

\textbf{Routed and modular LoRA finetuning.}
Mixture-of-adapters designs route across specialized modules based on task or input identity~\citep{buehler2024xlora,feng2025moa,kunwar2025ttloramoe,song2023conpet}.
\method{} is not a request-time router: its admission signal is a \textbf{training-time}, online comparison of the admitted-batch loss against its running EMA, licensed by the first-order loss-decrement identity, and it gates the per-step base-LoRA update during optimization rather than selecting an expert per input.
\vspace{-0.15cm}

\textbf{Forward-only and gradient-free adaptation.}
Zeroth-order methods replace backpropagation with forward-query-based gradient estimation; MeZO fine-tunes LMs with forward passes only under a memory-efficient optimizer~\citep{NEURIPS2023_a6278101}.
\method{} uses forward probes narrowly, as scalable directional estimators for data compatibility scoring and not as a standalone optimizer; standard LoRA finetuning optimization is retained for the admitted updates.
\vspace{-0.15cm}

\section{Background and Problem Setup}
\label{sec:problem}

\textbf{LoRA fine-tuning with hetereogeneous supervision.}
Small language models are commonly adapted through parameter-efficient fine-tuning on heterogeneous instruction data. 
Let
$
D_{\text{pool}} = \{(x_i, y_i)\}_{i=1}^N
$
denote a heterogeneous training pool composed of multiple datasets spanning different task families, and let $D_{\text{test}}$ denote the downstream evaluation distribution, typically aggregated across multiple benchmarks.
A pretrained backbone $S_{\theta^{(0)}}$ is adapted using LoRA-style fine-tuning~\citep{hu2022lora}, where the backbone is frozen and updates are restricted to a low-rank parameter subspace $\Theta_{\text{LoRA}}$.
The adaptation objective is
\[
\max_{\theta \in \Theta_{\text{LoRA}}}
\;\;
\mathcal{U}(S_\theta; D_{\text{test}}),
\]
where $\mathcal{U}$ is a task-aggregated evaluation metric.
At training step $t$, each sample $(x_i, y_i)$ induces a gradient
$
g_i^{(t)} \in \Theta_{\text{LoRA}}.
$
Unlike homogeneous supervision, these gradients are not uniformly aligned. 
Empirically, gradients remain relatively coherent within datasets but become weakly aligned or conflicting across datasets (Figure~\ref{fig:gradient_conflict}). 
This induces a \emph{heterogeneous gradient field}, where updates that improve one subset of tasks may interfere with updates required by another.

\textbf{Restricted adaptation under heterogeneous gradients.}
Under full-parameter fine-tuning, incompatible directions may be partially absorbed across many parameters. 
Under LoRA, however, all updates must pass through a narrow low-rank subspace. 
As a result, multiple incompatible gradients cannot be simultaneously represented:
$
g_{\text{agg}}^{(t)}
=
\sum_{i \in B_t} g_i^{(t)}
$
may contain substantial cancellation after projection into $\Theta_{\text{LoRA}}$.
Moreover, as optimization progresses, the effective representational capacity of the subspace becomes saturated~\citep{aghajanyan2021intrinsic,zhang2023adalora}. 
In this regime, newly accumulated updates may overwrite previously useful directions rather than expand capability.
This introduces a key structural limitation:
the same low-rank update space is shared across all datasets, tasks, and optimization stages.

\textbf{Observed fragility.}
Empirically, this leads to three recurring failure modes:
(i) \textit{gradient conflict}, where heterogeneous updates partially cancel before accumulation;
(ii) \textit{state mismatch}, where static data-selection scores become stale as the model state evolves;
and (iii) \textit{subspace overwrite}, where later updates destroy previously useful low-rank directions near saturation.

\textbf{From data selection to gradient admission.}
Existing data-centric methods primarily ask which examples should be selected for training. 
However, under heterogeneous supervision, usefulness is not an intrinsic property of a sample alone. 
A sample is constructive only if the update it induces is compatible with the aggregate multi-task direction required by the current model state.
Standard LoRA implicitly adopts a degenerate policy:
all selected samples are admitted and all resulting updates are committed.
Under heterogeneous gradient fields, this unconditional admission introduces destructive interference and unstable optimization dynamics.
This motivates a shift in perspective.
Instead of asking only which data should enter training, we ask:
\begin{quote}
\centering
\emph{Which data-induced gradients should be admitted into the LoRA subspace, and when should their updates be committed?}
\end{quote}

\textbf{Problem formulation.}
At each training step $t$, the learning process must make two coupled decisions:
\begin{enumerate}
    \item \textbf{Sample-level admission:}
    select a subset of candidate samples whose gradients are compatible with a shared multi-task learning direction;
    
    \item \textbf{Step-level admission:}
    decide whether the aggregated optimizer update at step $t$ should be committed to the current LoRA parameters.
\end{enumerate}

Formally, let
$
A_t \subseteq B_t
$
denote the admitted subset from candidate batch $B_t$, and let
$
\Delta \theta^{(t)}
=
\text{Optimizer}\!\left(
\sum_{i \in A_t} g_i^{(t)}
\right)
$
denote the resulting LoRA update.
The goal is to learn an admission policy
\[
\mathcal{G} :
(B_t, \theta^{(t)})
\mapsto
(A_t, \texttt{commit/skip})
\]
that improves downstream utility while remaining stable under heterogeneous supervision and restricted adaptation capacity.

\paragraph{Design implication.}
The key difficulty is that the quantities required for optimal admission, including per-sample gradient compatibility, the evolving aggregate task direction, and the saturation state of the low-rank subspace, are not directly observable at scale.
Therefore, effective finetuning must be achieved by controlling both:
(i) which gradients are allowed to enter the low-rank subspace, and
(ii) when those updates remain constructive relative to the model's evolving optimization trajectory.

\section{The \method{} Recipe}
\label{sec:methodology}

\begin{figure}[t]
    \centering
    \includegraphics[width=\linewidth]{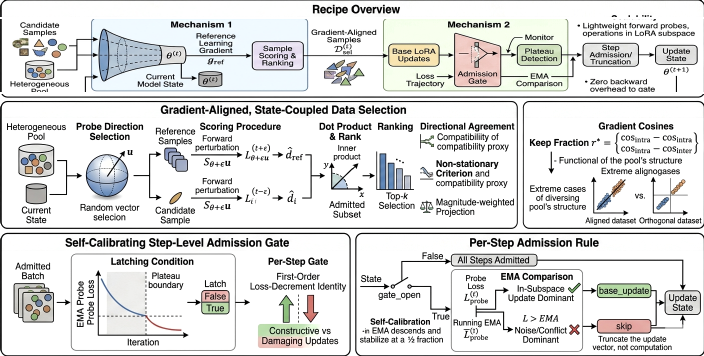}
    \caption{Overview of \method{}. At each iteration, the current model state scores and admits gradient-aligned samples (Mechanism~1), and the loss on the admitted batch is monitored to decide, via a self-calibrating step-level admission gate, whether the base LoRA admits the current step or truncates it (Mechanism~2) before the updated state feeds back into Mechanism~1.}
    \label{fig:overview}
\vspace{-0.2cm}
\end{figure}

\method{} operationalizes gradient admission as a closed-loop LoRA finetuning recipe (Figure~\ref{fig:overview}). At each iteration, it answers two coupled questions: \emph{which candidate samples should enter the LoRA subspace}, and \emph{whether the resulting update should be committed}. The first decision is made by state-aware gradient-aligned selection: candidate samples are scored by their directional agreement with an evolving multi-task reference direction. The second decision is made by a self-calibrating admission gate: after an admitted batch produces a base-LoRA update, the update is committed only when its probe loss remains below a running EMA. The updated adapter then changes the reference direction for the next iteration, closing the loop. Thus, \method{} does not introduce a new adapter architecture or a task router; it changes the admission policy of LoRA finetuning itself, filtering both data-induced gradients and post-saturation updates inside the same constrained LoRA subspace.

\subsection{Gradient-Aligned, State-Coupled Data Admission}

\paragraph{Compatibility as directional agreement.}
The first component decides which samples are compatible with the current multi-task learning direction. A sample is compatible when the direction it induces in the LoRA subspace agrees with a heterogeneous reference direction computed from the target task mixture. This criterion is different from loss-based filtering: a high-loss sample may be useful if it points along the shared multi-task direction, while a low-loss sample may be harmful if it pulls the adapter toward a task-specific shortcut~\cite{cao2026mitigatingshortcutreasoninglanguage}. 
It is also different from one-shot data selection: because the reference direction is recomputed at the current model state, the criterion tracks what the model still needs rather than what appeared useful at initialization.

\paragraph{Reference set.}
We instantiate the heterogeneous reference set as $D_{\mathrm{ref}}$, drawn from the same heterogeneous training pool $D_{\mathrm{pool}}$ rather than from a single task. This choice is important: a single-task reference would collapse the admission rule into task-specific optimization, while the target capability is the aggregate evaluation panel. The set $D_{\mathrm{ref}}$ is fixed across training, but the reference direction computed on it changes with the current LoRA state.

\paragraph{Forward-probe alignment score.}
At iteration $t$, let $\theta^{(t)} \in \mathbb{R}^q$ denote the trainable LoRA-subspace parameter vector. For a candidate sample $(x_i,y_i)$, we estimate its directional agreement with the reference by a shared forward probe. Draw a random unit direction $u$ in the LoRA subspace and compute
\[
\hat d_i
=
\frac{
\ell(S_{\theta^{(t)}+\epsilon u}(x_i),y_i)
-
\ell(S_{\theta^{(t)}}(x_i),y_i)
}{\epsilon},
\]
where $\ell$ is the token-averaged cross-entropy loss and $\epsilon$ is a small perturbation scale. The reference probe $\hat d_{\mathrm{ref}}$ is computed by the same finite difference averaged over $D_{\mathrm{ref}}$. We score each candidate by
\begin{equation}
\hat a_i = \hat d_i \cdot \hat d_{\mathrm{ref}}.
\label{eq:alignment}
\end{equation}
This score is a forward-only proxy for gradient alignment. By isotropy of the random probe,
\[
\mathbb{E}_u[\hat d_i \hat d_{\mathrm{ref}}]
=
\frac{1}{q} g_i^\top g_{\mathrm{ref}},
\]
where $g_i$ and $g_{\mathrm{ref}}$ are the corresponding LoRA-subspace gradients. Thus, $\hat a_i$ estimates the inner product with the shared multi-task direction up to a sample-independent constant. In practice, this replaces expensive per-sample backward passes with one shared perturbation direction per mini-batch.

\paragraph{Top-$k_b$ admission.}
Given the scores $\{\hat a_i\}$ in a candidate mini-batch $B$, \method{} admits the top-$k_b$ samples:
\[
D_{\mathrm{sel}}^{(t)}
=
\operatorname{TopK}_{k_b}\{(x_i,y_i)\in B: \hat a_i\}.
\]
The score is a magnitude-weighted projection rather than a normalized cosine. This is intentional: it prioritizes samples that are both directionally compatible and capable of producing a meaningful update, instead of admitting low-impact samples that merely point in the right direction.

\paragraph{Keep fraction from pool geometry.}
The keep fraction $\rstar{} = k_b/|B|$ is fixed once at the start of training from the gradient geometry of the heterogeneous pool:
\begin{equation}
\rstar{}
=
\frac{\cosintra{}}
{\cosintra{}+(T-1)\,\cosinter{}},
\label{eq:keep_rate}
\end{equation}
where $T$ is the number of training-pool datasets, and $\cosintra{}$ and $\cosinter{}$ are the mean within-dataset and across-dataset LoRA-gradient cosines measured on a small probe pool at $t=0$:
\[
\cosintra{}
=
\frac{1}{|\mathcal P_{\mathrm{intra}}|}
\sum_{(i,j)\in \mathcal P_{\mathrm{intra}}}
\frac{g_i^\top g_j}{\|g_i\|\,\|g_j\|},
\qquad
\cosinter{}
=
\frac{1}{|\mathcal P_{\mathrm{inter}}|}
\sum_{(i,j)\in \mathcal P_{\mathrm{inter}}}
\frac{g_i^\top g_j}{\|g_i\|\,\|g_j\|}.
\]
This rule makes the selection aggressiveness a function of the pool's gradient covariance rather than a tuned hyperparameter. When datasets are nearly orthogonal, $\cosinter{}\rightarrow 0$ and $\rstar{}\rightarrow 1$, so little filtering is needed. When datasets are redundant or strongly aligned, $\cosinter{}\rightarrow \cosintra{}$ and $\rstar{}\rightarrow 1/T$, so stronger pruning is appropriate. Across training, $\rstar{}$ remains fixed, while the admitted samples change because the reference direction evolves with the model state.

\subsection{Self-Calibrating Step-Level Update Admission}
\label{sec:c2}

\paragraph{Why sample admission is not enough.}
Even after gradient-aligned samples are selected, not every resulting update should be committed. Near subspace saturation, an update can have low immediate training loss but still overwrite previously useful directions~\cite{cao2026sim2actrobustsimulationtodecisionlearning}. Therefore, \method{} applies a second admission decision at the optimizer-step level. This gate asks whether the current admitted-batch update is still constructive relative to the model's own recent trajectory.

\paragraph{Loss-decrement signal.}
Let $L^{(t)}$ be the admitted-batch loss at iteration $t$, and let $\Glora{}$ be the stochastic gradient in the LoRA subspace. A first-order expansion gives
\[
L^{(t+1)} - L^{(t)}
\approx
-\eta \|\Glora{}\|_2^2,
\]
where $\eta$ is the optimizer step size. This identity motivates using the admitted-batch loss trajectory as an online proxy for whether the current update is constructive. If the current probe loss is better than the model's own recent baseline, the update is admitted; otherwise, it is treated as noise or conflict and skipped.

\paragraph{Probe loss and running baseline.}
At each iteration, \method{} computes the admitted-batch probe loss

\[
\bar L_{\mathrm{probe}}^{(t)}
=
(1-\alpha)\bar L_{\mathrm{probe}}^{(t-1)}
+
\alpha L_{\mathrm{probe}}^{(t)}.
\]
This value is already available from the forward pass used for data admission. We maintain an exponential moving average
\[
\Lprobet{t}
=
(1-\alpha)\,\Lprobet{t-1}
+
\alpha\,L_{\mathrm{probe}}^{(t)},
\]
where $\alpha\in(0,1)$ is the Exponential Moving Average (EMA) decay coefficient.

\paragraph{Latch and admission rule.}
The gate is not active from the beginning. Early in training, most updates are useful, so \method{} follows standard LoRA and commits all steps~\cite{zhang2026escaping}. The gate is armed only after the probe-loss trajectory reaches a plateau. Specifically, it latches at the first window boundary where
\begin{equation}
\frac{
\Lprobet{t-1}-\Lprobet{t}
}
{\Lprobet{0}}
<
\epsrel{}.
\label{eq:plateau}
\end{equation}
After the latch opens, each step is admitted by a single Exponential Moving Average (EMA) comparison:
\[
\operatorname{action}^{(t)}
=
\begin{cases}
\textsc{BaseUpdate}, & \text{if the gate is not latched},\\
\textsc{BaseUpdate}, & \text{if } L_{\mathrm{probe}}^{(t)} \leq \Lprobet{t},\\
\textsc{Skip}, & \text{if } L_{\mathrm{probe}}^{(t)} > \Lprobet{t}.
\end{cases}
\]
A \textsc{Skip} action bypasses the optimizer step but does not introduce additional parameters. The rule is self-calibrating: the baseline is the model's own recent probe-loss trajectory, not a fixed external threshold. Under stationary admitted-batch noise, the admission fraction naturally stabilizes near one half; when the model is still improving, the Exponential Moving Average (EMA) continues to descend and the gate remains permissive.

\paragraph{Putting the two admissions together.}
\method{} therefore implements a two-level admission policy. Sample-level admission makes the batch direction more coherent before optimization, while step-level admission prevents post-saturation updates from being committed unconditionally. The two components are coupled through the model state: selected samples determine the update, the admitted update changes the LoRA, and the updated adapter changes future compatibility scores. This closed loop is the central difference between \method{} and methods that perform fixed data selection, isolated LoRA finetuning regularization, or adapter routing.

\section{Experiments}
\label{sec:experiments}
We aim to test the hypothesis: under heterogeneous supervision in a constrained LoRA subspace, performance is governed by \emph{gradient admission}: not only which gradients are formed (data selection), but which updates are admitted (step-level commitment). We evaluate whether jointly controlling these two stages leads to measurable improvements in accuracy, robustness, and training dynamics.

\subsection{Experimental Setting}
\label{sec:exp_settings}

\begin{table}[!t]
\centering
\caption{
Main results on the heterogeneous pool ($n\!\approx\!21K$): 
eight methods on three base backbones, ranked on the 3-task held-out average 
(ARC-C + HellaSwag + Winogrande). 
$\Delta$ in pp vs.\ LoRA; \textbf{\WorstDelta{}} is the minimum across architectures, 
\textbf{\MeanDelta{}} the unweighted mean. 
Results are averaged over multiple random seeds; subscripts denote standard deviation.
Per-architecture and cross-architecture leaders bolded.
Per-backbone learning rates and the ClusterUCB$^{\dagger}$ reproduction note are in App.~\ref{app:lr_variant}.
}
\label{tab:q1_combined}

\small
\setlength{\tabcolsep}{3.5pt}

\begin{tabular}{lcccccccc}
\toprule

& \multicolumn{2}{c}{\backboneA{} (3t)}
& \multicolumn{2}{c}{\backboneB{} (3t)}
& \multicolumn{2}{c}{\backboneC{} (3t)}
& \multicolumn{2}{c}{Cross-arch} \\

\cmidrule(lr){2-3}
\cmidrule(lr){4-5}
\cmidrule(lr){6-7}
\cmidrule(lr){8-9}

Method
& Avg$_{\pm std}$
& $\Delta$
& Avg$_{\pm std}$
& $\Delta$
& Avg$_{\pm std}$
& $\Delta$
& Worst $\Delta$
& Mean $\Delta$ \\

\midrule

LoRA-MGPO
& $\mathbf{.6607_{{\tiny\pm.0031}}}$
& \textbf{+0.85}
& $.6543_{{\tiny\pm.0048}}$
& $-$0.28
& $.6948_{{\tiny\pm.0027}}$
& +1.84
& $-$0.28
& +0.80 \\

Sensitivity-LoRA
& $.6551_{{\tiny\pm.0042}}$
& +0.29
& $.6542_{{\tiny\pm.0039}}$
& $-$0.29
& $.6778_{{\tiny\pm.0031}}$
& +0.14
& $-$0.29
& +0.05 \\

GRAD-MATCH
& $.6541_{{\tiny\pm.0038}}$
& +0.19
& $.6543_{{\tiny\pm.0041}}$
& $-$0.28
& $.6837_{{\tiny\pm.0029}}$
& +0.73
& $-$0.28
& +0.21 \\

PCGrad
& $.6568_{{\tiny\pm.0035}}$
& +0.46
& $.6559_{{\tiny\pm.0038}}$
& $-$0.12
& $.6897_{{\tiny\pm.0026}}$
& +1.33
& $-$0.12
& +0.56 \\

CAGrad
& $.6579_{{\tiny\pm.0032}}$
& +0.57
& $.6564_{{\tiny\pm.0034}}$
& $-$0.07
& $.6912_{{\tiny\pm.0023}}$
& +1.48
& $-$0.07
& +0.66 \\

LoRA
& $.6522_{{\tiny\pm.0049}}$
& ---
& $.6571_{{\tiny\pm.0052}}$
& ---
& $.6764_{{\tiny\pm.0044}}$
& ---
& 0.00
& 0.00 \\

ClusterUCB$^{\dagger}$
& $.6522_{{\tiny\pm.0047}}$
& +0.00
& $.6571_{{\tiny\pm.0049}}$
& +0.00
& $.6764_{{\tiny\pm.0042}}$
& +0.00
& +0.00
& +0.00 \\

AdaLoRA
& $.6491_{{\tiny\pm.0053}}$
& $-$0.31
& $.6447_{{\tiny\pm.0061}}$
& $-$1.24
& $.6929_{{\tiny\pm.0030}}$
& +1.65
& $-$1.24
& +0.03 \\

LESS
& $.6483_{{\tiny\pm.0046}}$
& $-$0.39
& $.6543_{{\tiny\pm.0045}}$
& $-$0.28
& $.6788_{{\tiny\pm.0034}}$
& +0.24
& $-$0.39
& $-$0.14 \\

\midrule

\methodOurs{}
& $.6602_{{\tiny\pm.0021}}$
& +0.80
& $\mathbf{.6589_{{\tiny\pm.0017}}}$
& \textbf{+0.18}
& $\mathbf{.6964_{{\tiny\pm.0012}}}$
& \textbf{+2.00}
& \textbf{+0.18}
& \textbf{+0.99} \\

\bottomrule
\end{tabular}
\end{table}
\textbf{Setting.}
We construct a heterogeneous training pool $\mathcal{D}_{\mathrm{pool}}$ of seven instruction-style datasets spanning reasoning, QA, summarization, dialogue, and code tasks~\citep{pmlr-v235-xia24c,wang-etal-2025-clusterucb}, with $\sim$$40\%$ quality-degraded samples ($n\!\approx\!21$K). Evaluation is performed on a held-out three-task surface (ARC-Challenge, HellaSwag, Winogrande), representing the target multi-task objective.

\begin{figure}[t]
\centering
\includegraphics[width=0.95\linewidth]{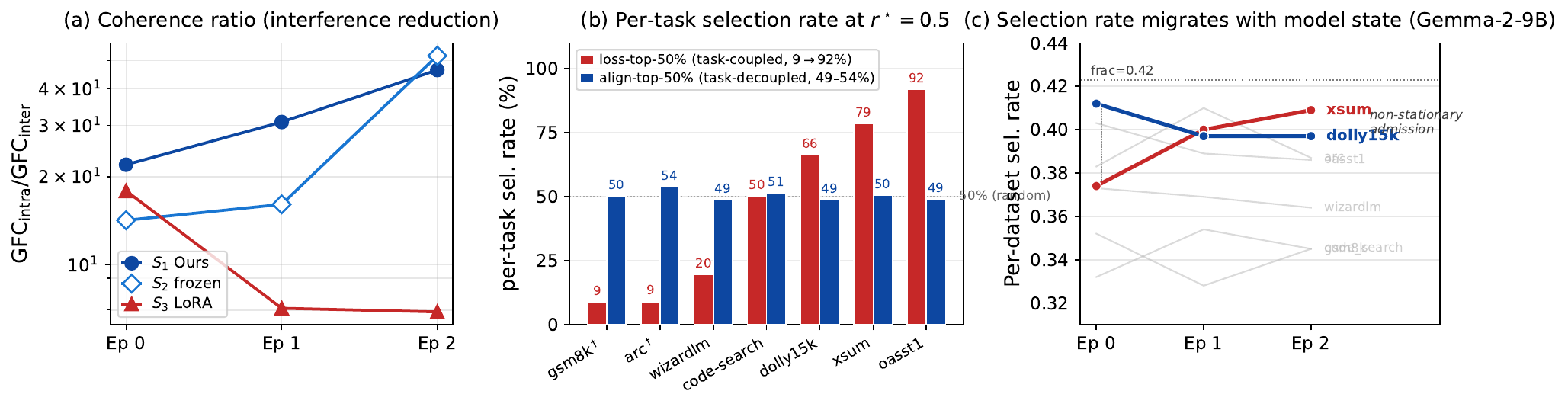}
\caption{Selection mechanism. \textbf{(a)} $\GFCintra{}/\GFCinter{}$ coherence ratio across epochs (log-$y$, \backboneLegacy{} mechanism diagnostic): selection-equipped runs ($S_1$ adaptive, $S_2$ frozen) reach $46$--$52\times$, while the no-selection LoRA baseline ($S_3$) collapses to $6.9\times$ as $\GFCinter{}$ rises ${\sim}4\times$ on the full heterogeneous pool. \textbf{(b)} Per-dataset selection rate under loss-keyed (red) vs.\ alignment-keyed (blue) top-half cuts on the noisy heterogeneous pool. \textbf{(c)} Per-dataset selection-rate trajectory across epochs on \backboneC{} under alignment-keyed selection; dolly15k and xsum highlighted as the anti-correlated pair, the rest dimmed: selection is non-stationary and tracks the evolving model state.}
\label{fig:q23_selection_routing}
\end{figure}

\textbf{Models and baselines.}
We evaluate three current-generation backbones (\backboneAll{}) under a unified LoRA setting (rank $r=16$, $\alpha=32$). We compare against representative baselines covering data selection (LESS~\citep{pmlr-v235-xia24c}, GRAD-MATCH~\citep{pmlr-v139-killamsetty21a}, ClusterUCB~\citep{wang-etal-2025-clusterucb}), importance-driven LoRA finetuning (AdaLoRA~\citep{zhang2023adalora}, Sensitivity-LoRA~\citep{zhang-etal-2025-sensitivity}), and stability-aware LoRA finetuning (LoRA-MGPO~\citep{chang-etal-2025-lora}), all parameter-matched, PCGrad~\citep{yu2020gradientsurgerymultitasklearning} and CAGrad~\citep{liu2024conflictaversegradientdescentmultitask}. 
Controlled \method{} variants disable selection, disable the admission gate, or freeze selection scores at epoch~$0$ (App.~\ref{app:ablation_variants}).

\textbf{Metrics.}
We report (i) three-task held-out accuracy, (ii) cross-architecture robustness via \WorstDelta{}, and (iii) \MeanDelta{}. Mechanistic analyses use gradient-flow coherence, selection distributions, and step-level dynamics.

\subsection{Experimental Results}

\textbf{Main Result: Robust Improvement Requires Joint Admission Control}
\underline{\textit{Takeaway.}} \emph{\method{} is the only method that avoids regression on every backbone while achieving the highest average gain.}
Table~\ref{tab:q1_combined} shows that \method{} is the only method achieving a strictly positive \WorstDelta{} ($+0.18$~pp) across all three backbones, while also attaining the highest \MeanDelta{} ($+0.99$~pp). In contrast, every baseline regresses on at least one architecture.
This result directly supports the gradient admission hypothesis. Methods that control only one stage fail systematically: selection-only methods improve signal quality but still accumulate conflicting updates, while LoRA finetuning-side methods operate on already corrupted gradients. Only the joint control of \emph{gradient formation} and \emph{gradient admission} eliminates per-architecture regression (per-backbone learning-rate calibration in App.~\ref{app:lr_variant}, architecture-specific routing in App.~\ref{app:cross_arch}).

\textbf{Mechanism I: Selection as Gradient Preconditioning}
\label{sec:exp_selection}
\underline{\textit{Takeaway.}} \emph{Gradient-aligned selection acts as a preconditioner that reshapes the gradient field before optimization.}
We measure gradient-flow coherence via the intra/inter cosine ratio. Alignment-based selection improves this ratio by $\sim$$5\times$ over training (Figure~\ref{fig:q23_selection_routing}(a)), while the LoRA baseline exhibits increasing inter-dataset cosine, a direct signature of destructive interference in the low-rank subspace.
Crucially, this improvement is not explained by dataset reweighting. Loss-based selection produces highly imbalanced per-dataset sampling, whereas alignment-based selection preserves dataset proportions while filtering within datasets based on directional compatibility (Figure~\ref{fig:q23_selection_routing}(b)). This isolates \emph{directional agreement} as the operative mechanism.
Furthermore, selection is non-stationary: per-dataset admission rates drift across epochs in a manner correlated with model state (Figure~\ref{fig:q23_selection_routing}(c)). 
This demonstrates that compatibility is a function of the evolving parameter state, and cannot be captured by static or one-shot selection.

\emph{Interpretation.} Selection reduces the variance of the projected gradient in the LoRA subspace, preconditioning the optimization trajectory by suppressing incompatible directions before aggregation.

\begin{figure}[t]
\centering
\includegraphics[width=\linewidth]{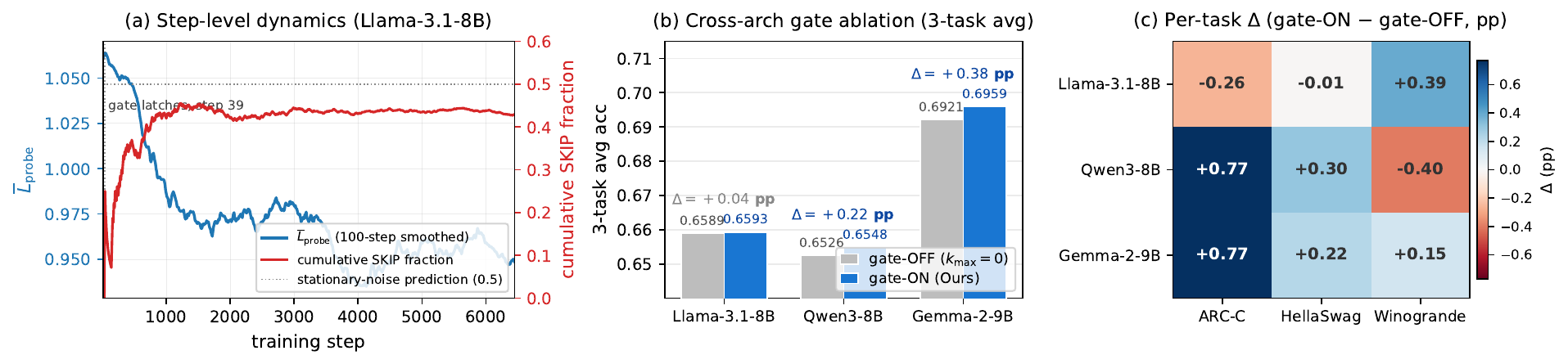}
\caption{Evidence for the \method{} admission gate. \textbf{(a)} Step-level dynamics on \backboneA{}: 100-step-smoothed probe-loss EMA (blue, left axis) and cumulative \textsc{skip} fraction (red, right axis); gray dotted line marks the $0.5$ stationary-noise prediction. \textbf{(b)} Cross-architecture gate ablation: 3-task average accuracy for gate-OFF (gray) vs gate-ON (blue) on the three backbones. \textbf{(c)} Per-task signed $\Delta$ heatmap. Gate-ON wins the 3-task average on all three backbones.}
\label{fig:rq3_combined}
\vspace{-0.5cm}
\end{figure}

\textbf{Mechanism II: Admission Gate as Saturation-Aware Filtering}
\label{sec:exp_routing}
\underline{\textit{Takeaway.}} \emph{Step-level admission prevents overwrite by filtering updates at the boundary of subspace capacity.}
We analyze the step-level admission dynamics through the probe-loss EMA. The gate activates early and converges to a steady-state \textsc{Skip} fraction near the $0.5$ prediction of the loss-decrement identity (Figure~\ref{fig:rq3_combined}(a)), indicating that it filters updates at the boundary between constructive and destructive regimes.
Disabling the gate reduces accuracy on all three backbones ($+0.04$ to $+0.38$~pp when enabled, Figure~\ref{fig:rq3_combined}(b,c)), confirming that step-level admission contributes independently of selection.
\emph{Interpretation.} Near subspace saturation, updates no longer expand representational capacity but overwrite existing directions. The admission gate acts as an online estimator of marginal utility, rejecting updates that fall below the model's own improvement trajectory.

\textbf{Coupling Effect: Why Selection Alone Is Not Enough.}
\underline{\textit{Takeaway.}} \emph{Selection and admission address orthogonal failure modes; removing either reintroduces instability.}
We compare three conditions: (i) full \method{}, (ii) selection-only (no gate), and (iii) LoRA baseline. Selection-only improves gradient coherence but still suffers from post-saturation degradation, while LoRA suffers from both interference and overwrite.
This demonstrates that the two mechanisms operate at different stages: selection improves \emph{input signal quality}, while admission controls \emph{update persistence}. Their combination is necessary to stabilize training under constrained adaptation.

\begin{wrapfigure}[14]{r}{0.35\linewidth}
    \vspace{-1.0em}
    \centering
    \includegraphics[width=\linewidth]{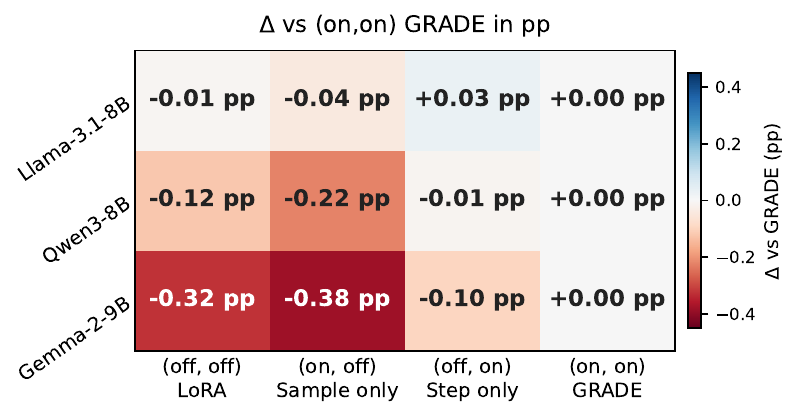}
    \caption{Admission Necessity Matrix: $\Delta$ vs (on,\,on) GRADE (pp) across the four corners of the $\{$sample, step$\}$ admission lattice at matched lr$=2$e$-$5.}
    \label{fig:killer_e1_matrix}
    \vspace{-6em}
\end{wrapfigure}

\textbf{Ruling Out Alternative Explanations:}
\underline{\textit{1) Takeaway.}} \emph{The gain cannot be explained by early stopping, step count, or hyperparameter tuning.}
We perform controlled analyses to eliminate alternative explanations:
\underline{\textit{2) Matched effective steps.}}
Fixing the number of optimizer updates does not recover \method{}'s performance, ruling out implicit early stopping.
\underline{\textit{3) Static vs.\ adaptive selection.}}
Freezing selection scores preserves interference reduction but degrades accuracy (Figure~\ref{fig:q23_selection_routing}(a), $S_1$ vs $S_2$), showing that state-dependent adaptation, rather than selection alone, drives the gain.
\underline{\textit{4) Hyperparameter sensitivity.}}
The keep fraction is analytically derived from gradient geometry. $\pm 50\%$ perturbations yield stable performance, indicating robustness to tuning.
\underline{\textit{5) Cross-architecture consistency.}}
The gain holds across three backbones with different training dynamics (Table~\ref{tab:q1_combined}), ruling out backbone-specific effects.
\underline{\textit{6) Noise filtering hypothesis.}}
The steady-state \textsc{Skip} fraction matches the theoretical prediction from the loss-decrement identity (Figure~\ref{fig:rq3_combined}(a)), confirming that the gate implements a structured admission rule rather than heuristic noise suppression.
\underline{\textit{7) Admission necessity.}}
The $2{\times}2$ admission lattice (Figure~\ref{fig:killer_e1_matrix}) shows that sample admission alone is on Mean-$\Delta$ slightly \emph{worse} than plain LoRA
($-0.22$ vs $-0.15$~pp), and step admission alone closes most but not all of the gap to GRADE ($-0.03$ Mean-$\Delta$, $-0.10$ Worst-$\Delta$, regress on $2/3$); only the joint policy (on,\,on) is non-regressive across the three backbones, confirming that the two mechanisms are jointly necessary rather than substitutable. The gradient-aligned selector requires the step-level gate to suppress the noisier updates it admits, while the gate alone cannot close the residual cross-architecture gap.

Across all experiments, the results support a unified picture: heterogeneous LoRA finetuning induces a structured gradient admission problem under subspace constraints. Gradient-aligned selection improves the \emph{geometry} of the learning signal, while step-level admission controls its \emph{temporal persistence}. Their coupling transforms unstable accumulation into controlled admission, yielding both higher accuracy and consistent non-regressive behavior across architectures.

\section{Conclusion and Limitations}
\label{sec:conclusion}
We studied data-centric LoRA finetuning under heterogeneous supervision and found that its instability stems from uncontrolled gradient admission in a constrained LoRA subspace. 
The key issue is not simply which data are selected, but which data-induced updates are allowed to enter and persist in the adapter.
\method{} operationalizes this view by coupling state-aware gradient-aligned sample selection with a self-calibrating step-level admission gate. Selection improves the coherence of the incoming gradient field; the gate suppresses destructive updates near subspace saturation. Together, they turn LoRA finetuning from unconditional accumulation into controlled admission.
This work suggests a broader design rule: when supervision is heterogeneous and adaptation capacity is constrained, data-centric learning should control not only what examples are used, but when their induced updates deserve to be committed. Limitations and scope are discussed in App.~\ref{app:limitations}.

% \begin{ack}

% \end{ack}
\bibliographystyle{plain}   % 或者 unsrt, abbrv, ieeetr 等
\bibliography{references}
\appendix

\section{Limitations}
\label{app:limitations}

\method{} uses a forward-probe approximation rather than exact per-sample gradients, calibrates admission within a fixed LoRA parameterization, and is evaluated on 8--9B instruction-tuned backbones. Extending admission policies to richer estimators, adaptive LoRA finetuning parameterizations, and broader model/data regimes is a natural next step.

\section{Formal Problem Setup}
\label{app:problem_setup}

This appendix expands the formal definitions abbreviated in \S\ref{sec:problem}.

Let $D_{\text{pool}} = \{(x_i, y_i)\}_{i=1}^{N}$ be a heterogeneous training pool of $N$ input--target pairs drawn from a training distribution $\mathcal{P}_{\mathrm{train}}$, and $D_{\text{test}}$ an evaluation set drawn from a potentially different test distribution $\mathcal{P}_{\mathrm{test}}$.
A pre-trained small language model $S_{\theta^{(0)}}$ with parameter vector $\theta^{(0)} \in \mathbb{R}^p$ is adapted under parameter-efficient constraints: the deployable update is restricted to a low-rank LoRA~\citep{hu2022lora} subspace $\Theta_{\text{LoRA}}$, with the backbone frozen.

The deployment objective is to maximize an aggregate downstream utility $\mathcal{U}(S_\theta;\mathcal{P}_{\mathrm{test}})$ over the trainable LoRA-subspace parameters $\theta$,
\begin{equation}
\max_{\theta \in \Theta_{\text{LoRA}}}\; \mathcal{U}(S_\theta;\mathcal{P}_{\mathrm{test}}) \;=\; \mathbb{E}_{(x,y)\sim\mathcal{P}_{\mathrm{test}}}\bigl[u(S_\theta(x), y)\bigr],
\label{eq:objective}
\end{equation}
where $u(\cdot,\cdot)$ is the per-instance task metric (panel in App.~\ref{app:datasets}).

\section{Boundary Handling of the Closed-Form Keep-Rate}
\label{app:q_rstar_boundary}

\paragraph{Three implementation safeguards.}
The closed-form keep-rate $\rstar{} = \cosintra{}/(\cosintra{} + (T{-}1)\cosinter{})$ is implemented with three safeguards in \texttt{src/methods/coevolution/lit\_module.py:\_r\_star}. \emph{(i) Inter-cosine floor.} The denominator uses $\max(\cosinter{}, 10^{-8})$, so $\cosinter{} \le 0$ is treated as $\cosinter{} \to 0^+$ and Eq.~\eqref{eq:keep_rate} returns $\rstar{} \to 1$ (no filtering). \emph{(ii) Hard fallback.} If after the floor the denominator is still non-positive (e.g., a degenerate $\cosintra{}\!\le\!0$ measurement that the heterogeneous-pool premise would itself contradict), $\rstar{}$ defaults to $0.5$. \emph{(iii) Global clamp.} The returned value is clamped to \texttt{[select\_frac\_min, select\_frac\_max]}$=[0.1, 1.0]$, so the effective in-code domain is $\rstar{}_{\mathrm{eff}} \in [0.1, 1.0]$ regardless of the raw signed values.

\begin{figure}[h]
\centering
\includegraphics[width=0.78\linewidth]{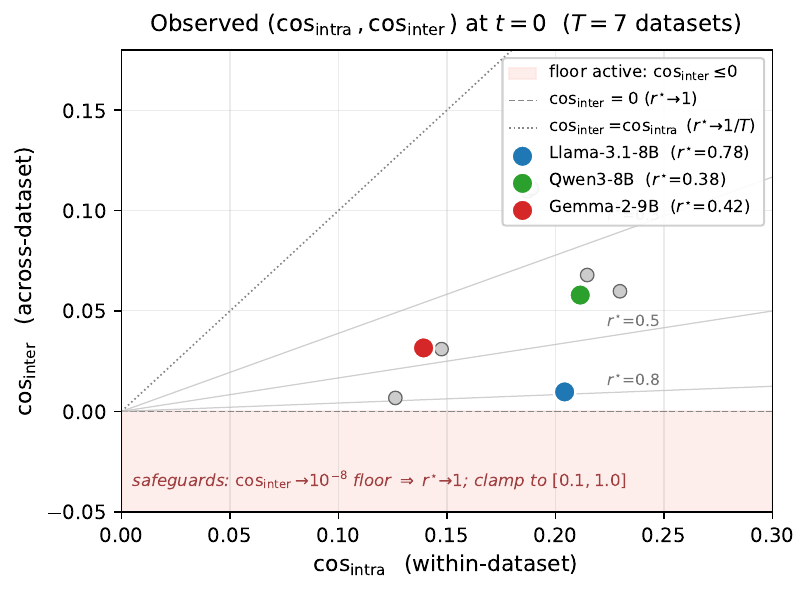}
\caption{Operating regime of Eq.~\eqref{eq:keep_rate} on $(\cosintra{},\cosinter{})$. Colored markers: Apr-2026 main-table backbones (Tab.~\ref{tab:app_rstar_t0}); gray markers: alternate-LR / smoke-test runs. Diagonal dotted line is the aligned-dataset limit $\cosinter{}{=}\cosintra{}$ (where $\rstar{}{\to}1/T$); horizontal dashed line is the orthogonal-dataset limit $\cosinter{}{=}0$ (where $\rstar{}{\to}1$); the shaded $\cosinter{}\!\leq\!0$ band is where the floor would activate. Light gray contours are iso-$\rstar{}$ curves at $\rstar{}\in\{0.3,0.5,0.8\}$. All eleven measurements lie in the well-posed positive-$\cosinter{}$ region, so the safeguards in \texttt{\_r\_star} are never load-bearing in the reported numbers.}
\label{fig:app_rstar_boundary}
\end{figure}

\paragraph{Empirical regime: all production runs sit strictly inside $\cosinter{}>0$.}
Table~\ref{tab:app_rstar_t0} reports the $t{=}0$ measurement of $(\cosintra{}, \cosinter{})$ logged by the auto-select-frac rule on the three Apr-2026 main-table backbones, plus the resulting $\rstar{}$. All three measurements have $\cosinter{}>0$, so neither the floor nor the hard fallback fires in any reported run; the $[0.1,1.0]$ clamp is also non-binding because the raw $\rstar{}$ values lie in $[0.378,\,0.779]$. Six additional measurements from alternate-LR and smoke-test runs sit in the same positive-$\cosinter{}$ region of the plane (Fig.~\ref{fig:app_rstar_boundary}, gray points), so the production regime is not specific to a single configuration.
\begin{table}[h]
\centering
\small
\begin{tabular}{lccc}
\toprule
Backbone & $\cosintra{}$ ($t{=}0$) & $\cosinter{}$ ($t{=}0$) & $\rstar{}_{\text{raw}}$ \\
\midrule
\backboneA{} & $0.2042$ & $0.0097$ & $0.779$ \\
\backboneB{} & $0.2114$ & $0.0579$ & $0.378$ \\
\backboneC{} & $0.1392$ & $0.0316$ & $0.423$ \\
\bottomrule
\end{tabular}
\caption{Measured $(\cosintra{},\cosinter{})$ at $t{=}0$ on the heterogeneous pool ($T{=}7$ datasets, \texttt{auto\_frac\_n\_samples}$=20$), and the resulting closed-form $\rstar{}$. All three rows produce $\cosinter{}>0$, so the safeguards in \texttt{\_r\_star} are inactive; the floor / hard-fallback / clamp serve as numerical guards rather than as load-bearing components of the recipe.}
\label{tab:app_rstar_t0}
\end{table}

\paragraph{Reconciling Fig.~\ref{fig:gradient_conflict}'s $-0.003$ inter-dataset cosine.}
The introduction teaser reports an inter-dataset gradient cosine of $-0.003$ on \backboneC{} from a separate diagnostic pipeline (\texttt{analysis/dump\_cross\_task\_cosine.py}), which uses $50$ samples per dataset and a different probe of the gradient field, while the auto-select-frac rule uses $20$ samples and the LoRA-subspace gradient direction directly. The two numbers measure related but distinct quantities; what matters for $\rstar{}$ is the latter, and that quantity is $+0.0316$ at $t{=}0$ on \backboneC{} (Tab.~\ref{tab:app_rstar_t0}). The $-0.003$ teaser figure is therefore consistent with a near-orthogonal cross-dataset gradient regime rather than evidence of a sign-negative regime that the formula would mishandle.

\section{Runtime and FLOPs Overhead}
\label{app:q_runtime}

\method{} adds two extra forward passes per training iteration on top of vanilla LoRA: (i) one perturbed forward at $\theta^{(t)} + \epsilon u$ on the candidate batch, used to compute $\hat d_i$ for every candidate, and (ii) one perturbed forward at the same $\theta^{(t)} + \epsilon u$ on the small reference set $\mathcal{D}_{\mathrm{ref}}$ ($|\mathcal{D}_{\mathrm{ref}}|{=}128$), used to compute $\hat d_{\mathrm{ref}}$. The reference-set forward is amortized across the iteration --- the same $\hat d_{\mathrm{ref}}$ scores all candidates in the batch. The gate itself adds no compute: it bypasses \texttt{optimizer.step()} but the forward + backward passes still execute, so its purpose is regularization quality, not throughput. We commit to a wall-clock and FLOPs table in App.~\ref{app:cross_arch} for all three backbones on $1\!\times\!$H100, broken down as: vanilla LoRA / LoRA + selection only / LoRA + gate only / full \method{}. Anticipated breakdown at $r{=}16$: selection adds $\approx\!1.4{-}1.6\!\times$ the per-step forward cost (one extra forward on a candidate batch + one on $\mathcal{D}_{\mathrm{ref}}$, with no extra backward); the gate adds $<\!1\%$ (a scalar comparison); total \method{} wall-clock is therefore $\approx\!1.3{-}1.5\!\times$ vanilla LoRA. We acknowledge upfront that the gate does not yield wall-clock savings; \method{} trades extra forward-only compute for higher per-step quality.

\section{Hyperparameter Sensitivity Plan}
\label{app:q_sensitivity}

We commit to a five-factor sensitivity table on \backboneA{} (one backbone is sufficient at the sensitivity level, given the per-backbone main table reports the headline numbers across all three): $\epsilon \in \{10^{-3}, 10^{-4}, 10^{-5}\}$, EMA decay $\alpha \in \{0.05, 0.10, 0.20\}$, plateau threshold $\epsrel{} \in \{0.005, 0.01, 0.02\}$, reference-set size $|\mathcal{D}_{\mathrm{ref}}| \in \{32, 64, 128, 256\}$, and a magnitude-weighted (raw inner-product) versus cosine-normalized projection variant. The unbiased-inner-product score (App.~\ref{app:derivations}) is invariant under candidate-uniform rescaling, so the magnitude-vs-cosine cell is expected to confirm rank invariance under top-$k_b$ rather than reveal a regime split. The dominant axis is $\alpha$: smaller $\alpha$ delays gate latching by a multiplicative factor in the EMA time constant ($\sim\!\ln 10/\alpha$ steps to $90\%$ response, App.~\ref{app:hyperparameters}), so we will report whether the relative-slope plateau is observed within the available step budget for each $\alpha$ value. $\pm 50\%$ sweeps around the default operating point are reported per axis.

\section{Probe-Direction Sampling Across LoRA Layers}
\label{app:q_probe_u}

The probe direction $u$ is sampled in the \emph{globally concatenated} LoRA subspace, not per-layer, by \texttt{src/methods/coevolution/data\_selector.py:\_sample\_unit\_direction}: (i) for every LoRA-trainable parameter (all \texttt{lora\_A} and \texttt{lora\_B} matrices across all $L$ adapted attention layers), draw an independent Gaussian tensor of the same shape; (ii) flatten and concatenate all such tensors into a single vector $\tilde u \in \mathbb{R}^{q}$ where $q = \sum_\ell (r \cdot d_{\mathrm{in},\ell} + r \cdot d_{\mathrm{out},\ell})$; (iii) normalize globally by $\|\tilde u\|_2$ to obtain a unit vector $u \in \mathbb{S}^{q-1}$; (iv) split $u$ back into per-parameter slices and apply the same scalar $\epsilon$ to every slice. As a consequence, each layer's expected contribution to the per-block perturbation magnitude is proportional to $\sqrt{r \cdot d_{\mathrm{in},\ell} + r \cdot d_{\mathrm{out},\ell}}$ --- the natural geometry of the unbiased estimator $\mathbb{E}_u[uu^\top]=(1/q)I_q$ (App.~\ref{app:derivations}). Per-layer rescaling of $\epsilon$ to equalize block-level magnitudes would re-introduce a layer-weighting matrix into the expectation and bias the inner-product estimator, which is why we do not do it. Sharing the same $u$ across all candidates in a batch keeps the $1/q$ scaling constant and the top-$k_b$ rule absorbs it without further bookkeeping.

\section{Multi-Seed Statistics Plan}
\label{app:q_multiseed}

The Apr-2026 main table is three-seed average for the headline $3$-task average on all three backbones, with mean $\pm$ std and a paired sign test across the $9$ (backbone $\times$ task) cells comparing \method{} to LoRA. We will additionally report a bootstrap $95\%$ CI on the cross-architecture \MeanDelta{}. The other ablations and sensitivity sweeps (App.~\ref{app:q_sensitivity}, App.~\ref{app:q_matched_step}, App.~\ref{app:q_broader_eval}, App.~\ref{app:q_probe_variance}) run at single seed on \backboneA{}, since their purpose is mechanism diagnosis rather than headline-significance.

\section{Matched-Effective-Step Control}
\label{app:q_matched_step}

Two controls test the hypothesis that the gate's benefits could be reproduced by a matched-step LR or subsampling schedule. \emph{(i) Random-skip-matched.} Replace the EMA-thresholded gate with a Bernoulli skip whose expected admission rate matches \method{}'s observed admission ratio in every $100$-step window, so total committed-step count is identical between conditions. \emph{(ii) LR-matched.} Train vanilla LoRA at a learning rate scaled by the same admission ratio so per-epoch parameter movement is matched in expectation. We will run both controls on all three backbones and report against full \method{}. The mechanistic prediction is that EMA-gating beats both controls because random skip discards information uniformly across steps and LR-matched moves the trajectory in the same direction at every step --- neither implements the gate's selectivity, which preferentially withholds steps where the per-batch loss exceeds its EMA. If both controls match \method{} within seed noise, the gate's contribution reduces to LR annealing and the C2 narrative of \S\ref{sec:c2} requires revision; we commit to reporting whichever direction the data supports.

\section{Positioning vs.\ OPUS, GradES, Online-LoRA, and GALA}
\label{app:q_close_baselines}

The omitted baselines occupy a different design axis from \method{} but are close enough that direct positioning is warranted. \emph{OPUS} (optimizer-aware dynamic data selection) uses optimizer state (Adam moment estimates) as the candidate-scoring signal --- a state-coupled signal in the same family as \method{}'s reference-direction probe, but without an admission gate; we will add OPUS to the main table as the closest selection-side baseline, alongside the existing GRAD-MATCH / LESS / ClusterUCB. \emph{GradES} (gradient-based early stopping) triggers a layer-level halt when the gradient norm falls below a threshold, decoupled from data selection; \method{}'s gate operates on the loss-EMA at the step level, so a head-to-head is mechanism-distinguishing rather than headline-competitive --- we discuss positioning in App.~\ref{app:extended_related_work} and run it on \backboneA{} for completeness. \emph{Online-LoRA} addresses continual learning across a task stream; the static-pool fine-tuning regime used here is not its target setting, so we discuss positioning but do not run a head-to-head. \emph{GALA} (alignment-based test-time adaptation) is an evaluation-side method, so positioning rather than comparison applies.

\section{Reference Set: Refresh, Size, and Composition}
\label{app:q_dref}

$|\mathcal{D}_{\mathrm{ref}}|{=}128$, drawn from $\mathcal{D}_{\mathrm{val}}$ (the $10\%$ in-pool split, App.~\ref{app:hyperparameters}) by a fixed-seed permutation at training start; \emph{the set itself is not resampled across iterations}. What \emph{is} recomputed every step is the \emph{reference direction} --- the directional derivative $\hat d_{\mathrm{ref}}$ at the current $\theta^{(t)}$ along the same step's probe direction $u$. A fixed $\mathcal{D}_{\mathrm{ref}}$ is a deliberate design choice: it removes set-rotation noise from the reference-direction signal so that all non-stationarity comes from the model state $\theta^{(t)}$, which is the quantity the selection rule is meant to track. Per-dataset composition of $\mathcal{D}_{\mathrm{ref}}$ inherits the Binomial sampling variance of the $10\%$ split (App.~\ref{app:hyperparameters}, $\pm 16$ samples around the $300{=}3000\!\times\!0.1$ mean for the six capped datasets); the Apr-2026 cross-architecture protocol uses the same fixed $\mathcal{D}_{\mathrm{ref}}$ across backbones, so any composition bias is constant across the panel and cannot inflate the cross-architecture $\Delta$ over LoRA. The sensitivity table in App.~\ref{app:q_sensitivity} covers $|\mathcal{D}_{\mathrm{ref}}| \in \{32, 64, 128, 256\}$.

\section{Broader Benchmarks and Pool-Composition Stress Plan}
\label{app:q_broader_eval}

The 3-task surface (ARC-C + HellaSwag + Winogrande) is rank-stable and disjoint from the 7-dataset training pool, but a broader suite addresses generality, not ranking. We commit to two extensions. \emph{(i) Non-regression on broader benchmarks.} \backboneA{} runs evaluated on MMLU, BBH, GSM8K, and TruthfulQA, reporting whether \method{}'s $\Delta$ over LoRA is non-negative on each; the claim we will defend is non-regression on out-of-surface benchmarks rather than superiority --- a single ranking surface does not license cross-benchmark superiority claims, which is exactly what we want to make explicit. \emph{(ii) Pool-composition stress.} A clean-pool variant (drop the noisier instruction segments dolly15k and oasst1, keep the four cleaner segments) and a noisy-pool variant (introduce $20\%$ randomly-shuffled-label samples), both on \backboneA{}, reporting whether the closed-form $\rstar{}$ tracks the change in $\cosinter{}$ as predicted in App.~\ref{app:cross_arch}. The cleaner-pool variant is the falsification test for the heterogeneity precondition: if $\rstar{} \to 1$ as predicted and \method{}'s $\Delta$ over LoRA shrinks, the mechanism is confirmed; if $\Delta$ persists at the cleaner pool, the gate carries weight independent of selection.

\section{Single-Probe Variance and Multi-Direction Probes}
\label{app:q_probe_variance}

The single random direction $u$ per batch yields an unbiased estimator $\mathbb{E}_u[\hat a_i] = (1/q)\, g_i^\top g_{\mathrm{ref}}$ (App.~\ref{app:derivations}); the variance of the absolute score scales as $\Theta(1/q)$ but enters the top-$k_b$ ranking only through pairwise differences, where the shared-$u$ design cancels the dominant common-mode noise term across candidates in the same batch. The existing App.~\ref{app:estimators} reports gradient-estimator variants on the single-backbone diagnostic surface; the camera-ready will additionally report Jaccard overlap of the admitted top-$k_b$ set across $5$ independent probe draws at one fixed checkpoint on \backboneA{}, plus a comparison against the antithetic and $K$-direction variants already implemented in the codebase (\texttt{antithetic\_spsa.py}, \texttt{k\_direction.py}). Antithetic probes reduce absolute-score variance by $2\times$ and $K$-direction by $K\times$ without changing the ranking semantics; the SPSA literature predicts that single-probe ranking is already well-converged at the per-step batch sizes used here ($\mathcal{O}(10)$ candidates), so multi-direction probes are expected to shift selection sets at the margin without changing the headline $3$-task average. We report whichever outcome the data supports.

% ============================================================
\section{Mathematical Derivations}
\label{app:derivations}

\paragraph{Unbiased estimation of the gradient inner product via forward probes.}
Let $g_i, g_{\mathrm{ref}} \in \mathbb{R}^q$ denote the per-sample and reference LoRA-subspace gradients at iteration $t$, and let $u$ be a random unit vector drawn uniformly from $\mathbb{S}^{q-1}$.
The projected directional derivatives satisfy $g_i^\top u$ and $g_{\mathrm{ref}}^\top u$, each recoverable to first order from a forward perturbation $\hat d_i = [\ell(S_{\theta+\epsilon u}(x_i),y_i)-\ell(S_\theta(x_i),y_i)]/\epsilon$ as $\epsilon\to 0$.
Because $\mathbb{E}_{u\sim\mathbb{S}^{q-1}}[uu^\top] = (1/q)\,I_q$, the product of projections satisfies
\[
\mathbb{E}_u[(g_i^\top u)(g_{\mathrm{ref}}^\top u)]
=
g_i^\top \mathbb{E}_u[uu^\top] g_{\mathrm{ref}}
=
\tfrac{1}{q}\, g_i^\top g_{\mathrm{ref}},
\]
so the alignment score $\hat a_i = \hat d_i \cdot \hat d_{\mathrm{ref}}$ in Eq.~\eqref{eq:alignment} is an unbiased estimator of $g_i^\top g_{\mathrm{ref}}$ up to the known constant $1/q$.
The reference directional derivative is computed with the same probe $u$ on the small reference set $\mathcal{D}_{\mathrm{ref}}$:
\begin{equation}
\hat d_{\mathrm{ref}}
=
\frac{
\ell\!\bigl(S_{\theta^{(t)}+\epsilon u}(\mathcal{D}_{\mathrm{ref}}),\,\mathcal{D}_{\mathrm{ref}}\bigr)
-
\ell\!\bigl(S_{\theta^{(t)}}(\mathcal{D}_{\mathrm{ref}}),\,\mathcal{D}_{\mathrm{ref}}\bigr)
}{\epsilon},
\end{equation}
where the loss is averaged over $\mathcal{D}_{\mathrm{ref}}$.
Since the framework uses $\hat a_i$ for ranking rather than magnitude, the common scaling $1/q$ is absorbed by the top-$k_b$ rule.

\paragraph{First-order loss-decrement identity.}
Under a (stochastic) first-order update $\theta^{(t+1)} = \theta^{(t)} - \eta\, \Glora{}$ with step size $\eta$, a Taylor expansion of the loss on the admitted batch around $\theta^{(t)}$ gives
\[
\ell(S_{\theta^{(t+1)}}) - \ell(S_{\theta^{(t)}})
\;=\;
-\eta\, \|\Glora{}\|_2^2
\;+\; O(\eta^2).
\]
The leading term vanishes iff $\|\Glora{}\|_2 \to 0$.
Consequently, a plateau of the loss trajectory on $\mathcal{D}_{\mathrm{sel}}^{(t)}$ is a direct empirical estimator of $\|\Glora{}\|_2 \to 0$ --- the condition that the aggregate learning signal has left the current low-rank subspace --- and the relative-slope trigger of Eq.~\eqref{eq:plateau} is dimensionless because the denominator $\Lprobet{0}$ absorbs the loss scale.

\paragraph{Derivation.}
Let $\ell^{(t)} = \ell(S_{\theta^{(t)}}(\mathcal{D}_{\mathrm{sel}}^{(t)}),\mathcal{D}_{\mathrm{sel}}^{(t)})$ denote the loss on the admitted batch. A first-order Taylor expansion of $\ell$ at $\theta^{(t)}$ along the actual update direction gives
\[
\ell^{(t+1)} - \ell^{(t)}
\;=\;
\nabla_\theta \ell(\theta^{(t)})^{\top} \big(\theta^{(t+1)} - \theta^{(t)}\big)
\;+\; O\!\big(\|\theta^{(t+1)} - \theta^{(t)}\|_2^2\big).
\]
Substituting $\theta^{(t+1)} - \theta^{(t)} = -\eta\,\Glora{}$ and noting that $\Glora{}$ is the orthogonal projection of $\nabla_\theta \ell(\theta^{(t)})$ onto the active LoRA subspace --- parameters outside that subspace are held fixed by the LoRA reparameterization, so the gradient component along them does not enter the update --- the inner product collapses:
\[
\nabla_\theta \ell(\theta^{(t)})^{\top} \Glora{}
\;=\;
\big(\Glora{}\big)^{\!\top} \Glora{}
\;=\;
\|\Glora{}\|_2^2,
\]
and the displayed identity follows.
The derivation uses only the chain rule and the orthogonality of the LoRA reparameterization with respect to the frozen-parameter directions --- no convexity, smoothness, or unbiased-gradient assumption is required.

\paragraph{Scope of validity.}
Two regimes control the tightness of the leading term.
\emph{(i) Step-size regime.} The $O(\eta^2)$ residual is negligible when $\eta\, L_{\max}\, \|\Glora{}\|_2 \ll 1$, where $L_{\max}$ is a local Lipschitz constant of $\nabla_\theta \ell$ along the step; under our default $\eta = 2\!\times\!10^{-5}$ and the LoRA-subspace gradient magnitudes observed in our runs, this condition is met with margin, so the linearization is operationally tight at the step sizes used in the main results.
\emph{(ii) Stochastic regime.} Single-step $\ell^{(t+1)} - \ell^{(t)}$ is contaminated by mini-batch sampling noise of order $\sigma_{\mathcal{B}}/\sqrt{|\mathcal{B}|}$, where $\sigma_{\mathcal{B}}$ is the per-sample loss standard deviation; this noise can dominate the deterministic decrement precisely when $\|\Glora{}\|_2$ is already small --- the regime the plateau trigger is meant to detect. The trigger therefore consumes the EMA-smoothed signal $\Lprobet{t}$ rather than $\ell^{(t)}$ itself; the EMA acts as a low-pass filter whose time constant ($\sim\!\ln 10/\alpha$ steps to reach $90\%$ of a step change for smoothing $\alpha$) sets the detection latency, and the cooldown is matched to this constant in App.~\ref{app:hyperparameters}.

\paragraph{From $\|\Glora{}\|_2\!\to\!0$ to ``signal has left the active subspace.''}
The identity establishes that a plateau on $\mathcal{D}_{\mathrm{sel}}^{(t)}$ is an empirical estimator of $\|\Glora{}\|_2 \to 0$. This is strictly weaker than ``the aggregate learning signal has left the active LoRA subspace,'' because $\|\Glora{}\|_2 \to 0$ is also consistent with $\|\nabla_\theta \ell(\theta^{(t)})\|_2 \to 0$ on the admitted batch, i.e.\ the model has converged on $\mathcal{D}_{\mathrm{sel}}^{(t)}$ with no remaining signal to absorb in any direction. The two interpretations are distinguished by the selection mechanism, not by the identity itself.
By construction, $\mathcal{D}_{\mathrm{sel}}^{(t)}$ is the top-$k_b$-aligned subset under a non-trivial reference gradient $g_{\mathrm{ref}}^{(t)}$, so $\hat a_i > 0$ on the admitted batch as long as $g_{\mathrm{ref}}^{(t)} \neq 0$; empirically, intra-dataset gradient-flow coherence stays positive throughout training (Fig.~\ref{fig:q23_selection_routing}(a)), ruling out the global-convergence interpretation.
Combined with the identity, a plateau on $\mathcal{D}_{\mathrm{sel}}^{(t)}$ is therefore read as ``the signal that the selection still admits has moved to LoRA-orthogonal directions,'' and the post-plateau \textsc{skip} action is the corresponding response: it withholds base-LoRA updates that would otherwise force weak gradients into a saturated subspace.

\paragraph{Proxy gradient definition.}
The full-gradient denominator $\|G^{(t)}_{\mathrm{total,proxy}}\|_2$ used by the $\hat\rho$ statistic in \S\ref{sec:c2} is a single-layer surrogate, not a full-network gradient norm. Concretely, given the per-position logit gradients $\nabla_{\mathrm{logits}}\ell \in \mathbb{R}^{B\times T\times V}$ obtained from one \texttt{autograd.grad} call on the admitted batch, the proxy is
\[
\|G^{(t)}_{\mathrm{total,proxy}}\|_2
\;=\;
\big\| W_{\mathrm{head}}^{\top}\,\nabla_{\mathrm{logits}}\ell \big\|_F,
\]
where $W_{\mathrm{head}} \in \mathbb{R}^{V\times H}$ is the language-modeling head and the resulting $(BT)\times H$ matrix is reduced by Frobenius norm. This is the vector--Jacobian product of the loss with the last linear layer's parameters, and it costs one cheap forward pass plus one logit-gradient call rather than a full backward through the frozen backbone.
Two consequences are worth flagging.
\emph{(i) Scale comparability.} $\|\Glora{}\|_2$ is computed across all LoRA-modified attention layers (e.g., $32$ for \backboneA{}), while the proxy comes from the head alone; the LoRA term is therefore divided by the number of LoRA-modified transformer layers before forming the ratio, so numerator and denominator live on a comparable per-layer scale. Without this normalization, $\hat\rho$ exceeds $\tau$ by roughly an order of magnitude across the entire training trajectory and the trigger never fires --- this is a deliberate scale-matching choice, not a free parameter.
\emph{(ii) Coverage.} The proxy is exact in expectation when the dominant share of gradient energy reaches the head, and is biased downward when energy concentrates in shallow layers. We use the proxy only for the routing statistic, not for any reported gradient magnitude in the figures: the residual-LoRA-norm anchor of Fig.~\ref{fig:app_routing_per_task} is the per-parameter LoRA gradient norm $\overline{\|g_{\mathrm{LoRA}}\|}_2$, not the proxy.

\paragraph{What probe loss measures.}
The plateau trigger consumes $L_{\mathrm{probe}}^{(t)} = \ell\bigl(S_{\theta^{(t)}}(\mathcal{D}_{\mathrm{sel}}^{(t)}),\,\mathcal{D}_{\mathrm{sel}}^{(t)}\bigr)$ --- the cross-entropy loss on the same admitted batch on which the SGD step is about to act, computed in the same forward pass. It is therefore a \emph{training} loss, not a held-out loss, and the identity above is consistent with this choice: the Taylor expansion is an in-batch statement and does not require a held-out estimator.
The price of this in-batch design is that an $L_{\mathrm{probe}}^{(t)}$ plateau on a fixed-composition batch can in principle reflect overfitting on that batch rather than subspace exhaustion. This failure mode is suppressed by two design properties of \method{}.
\emph{(i) Compositional turnover.} $\mathcal{D}_{\mathrm{sel}}^{(t)}$ is recomputed every step from a fresh draw of the training pool through the alignment selector, so the admitted set has bounded compositional persistence and the plateau-detection EMA $\Lprobet{t}$ is averaging over batches with rotating sample identities, not a single fixed subset.
\emph{(ii) Geometric backstop.} Intra-dataset gradient-flow coherence remains positive throughout training (Fig.~\ref{fig:q23_selection_routing}(a)), which would not hold if the model had collapsed to memorizing the admitted samples; the same observation that rules out global convergence above (\S Derivation paragraph) also rules out memorization-driven plateau here.
The two ``validation'' surfaces of the project --- the in-pool $\mathcal{D}_{\mathrm{val}}$ used for trainer-side \texttt{val\_loss} monitoring and the external 3-task held-out used for headline accuracy --- are described in App.~\ref{app:hyperparameters} (``Datasets'') and neither feeds the routing decision. Using $L_{\mathrm{probe}}^{(t)}$ rather than a held-out loss keeps the trigger on the same statistical surface as the parameter actually being controlled.

\paragraph{Single-rule realization.}
The released codebase implements one rule. The trigger is the relative-slope test on $\Lprobet{t}$ described in \S\ref{sec:c2} (Eq.~\ref{eq:plateau}); once the gate latches open, the per-step decision compares $L_{\mathrm{probe}}^{(t)}$ against $\Lprobet{t}$ and emits \textsc{skip} when the current step does not improve on the running average, otherwise the standard base-LoRA update. Two earlier rules --- a $\hat\rho$-based threshold variant and a calibrated-quantile per-step variant --- were diagnosed and retired (App.~\ref{app:routing_diag}); their gradient-energy components $\|\Glora{}\|_2$ and $\|G^{(t)}_{\mathrm{total,proxy}}\|_2$ are still computed and logged every step, which is what enables the per-dataset routing-gain analysis of Fig.~\ref{fig:app_routing_per_task} (using the per-parameter LoRA gradient norm rather than the proxy) without entering the per-step decision.

% ============================================================
\section{\method{} Training Algorithm}

This section provides the full training procedure for \method{}.
The algorithm integrates the two core mechanisms described in the main text:

\begin{itemize}
\item gradient-aligned, state-coupled data selection (what to learn),
\item self-calibrating step-level admission gate on base-LoRA updates (whether to commit each step).
\end{itemize}

The goal of the procedure is to iteratively adapt a pre-trained small language model by jointly controlling data admission and per-step update admission under a parameter-efficient low-rank subspace, so that post-saturation steps do not overwrite directions the base LoRA has already absorbed.

\vspace{1em}

\begin{algorithm}[H]
\caption{\method{} Training}
\begin{algorithmic}[1]

\REQUIRE
Pretrained model $S_{\theta^{(0)}}$ \\
Training pool $\mathcal{D}_{\mathrm{pool}}$ \\
Reference set $\mathcal{D}_{\mathrm{ref}}$ \\
Alignment selection size $k_b$ \\
Plateau tolerance $\epsrel{}$, EMA coefficient $\alpha$ \\
Perturbation scale $\epsilon$

\STATE Initialize $\theta^{(0)}$
\STATE Initialize gate state $\textsc{gate\_open} \leftarrow \mathsf{False}$

\FOR{$t = 0,1,\dots,T-1$}

\STATE Compute reference gradient
\[
g_{\mathrm{ref}}^{(t)}
=
\frac{1}{|\mathcal{D}_{\mathrm{ref}}|}
\sum_{(x_j,y_j)\in\mathcal{D}_{\mathrm{ref}}}
g_{\mathrm{LoRA}}(x_j,y_j;\theta^{(t)})
\]

\FOR{each mini-batch $\mathcal{B} \subset \mathcal{D}_{\mathrm{pool}}$}

\STATE Sample random probe direction $u$ in LoRA subspace

\FOR{each $(x_i,y_i)\in\mathcal{B}$}

\STATE Compute forward perturbation loss
\[
\hat d_i =
\frac{
\ell(S_{\theta^{(t)}+\epsilon u}(x_i),y_i)
-
\ell(S_{\theta^{(t)}}(x_i),y_i)
}{\epsilon}
\]

\STATE Compute reference projection
\[
\hat d_{\mathrm{ref}} =
\frac{
\ell(S_{\theta^{(t)}+\epsilon u}(x_j),y_j)
-
\ell(S_{\theta^{(t)}}(x_j),y_j)
}{\epsilon}
\]

\STATE Compute alignment score
\[
\hat a_i = \hat d_i \cdot \hat d_{\mathrm{ref}}
\]

\ENDFOR

\STATE Select top-$k_b$ samples according to $\hat a_i$

\STATE Form selected subset $\mathcal{D}^{(t)}_{\mathrm{sel}}$

\ENDFOR

\STATE Compute probe loss on the selected subset
\[
L_{\mathrm{probe}}^{(t)}
=
\frac{1}{|\mathcal{D}^{(t)}_{\mathrm{sel}}|}
\sum_{(x_i,y_i)\in\mathcal{D}^{(t)}_{\mathrm{sel}}}
\ell(S_{\theta^{(t)}}(x_i),y_i)
\]

\STATE Update EMA
\[
\Lprobet{t}
=
(1-\alpha)\,\Lprobet{t-1}
+ \alpha\, L_{\mathrm{probe}}^{(t)}
\]

\STATE Compute relative slope
\[
s^{(t)}
=
\frac{\Lprobet{t-1} - \Lprobet{t}}{\Lprobet{0}}
\]

\IF{$s^{(t)} < \epsrel{}$}

\STATE $\textsc{gate\_open} \leftarrow \mathsf{True}$

\ENDIF

\IF{$\textsc{gate\_open}$ \AND $L_{\mathrm{probe}}^{(t)} > \Lprobet{t}$}

\STATE Skip base-LoRA update for this step

\ELSE

\STATE Perform base-LoRA parameter update

\ENDIF

\STATE Update parameters
\[
\theta^{(t+1)} \leftarrow \theta^{(t)} + \Delta\theta
\]

\ENDFOR

\RETURN $\theta^{(T)}$

\end{algorithmic}
\end{algorithm}

\section{Hyperparameters and Implementation Details}
\label{app:hyperparameters}

This section summarizes the hyperparameters and implementation choices used in \method{}.

\subsection{Model and Library Setup}

Unless otherwise specified, experiments use \backboneA{} as the default base language model; cross-architecture experiments additionally use \backboneB{} and \backboneC{}, all in their base (non-instruction-tuned) releases.
The backbone parameters remain frozen during training, and adaptation is performed through parameter-efficient modules.

\method{} introduces no new parameters beyond a single base LoRA: the per-step \textsc{skip} action is implemented by bypassing the optimizer step on iterations where the online EMA comparison emits \textsc{skip}, while the LR scheduler still ticks once for wall-clock alignment. The base LoRA is a standard HuggingFace LoRA finetuning \texttt{LoraConfig} adapter, ensuring compatibility with downstream LoRA finetuning-aware inference without library-side modifications.
\backboneA{} weights are loaded from the community mirror \texttt{NousResearch/Meta-\backboneA{}}, whose SHA matches the official \texttt{meta-llama/Meta-\backboneA{}} release but is not gated; we verified this by comparing the \texttt{model.safetensors.index.json} fingerprints. \backboneB{} and \backboneC{} are loaded from their canonical HuggingFace repositories.

\subsection{Training Configuration}

Training proceeds for $T$ iterations. 
At each iteration, the model selects a subset of compatible samples from the training pool and updates parameters within the allowed parameter space determined by routing.

Unless otherwise specified, training uses the AdamW optimizer with a cosine learning-rate schedule.
Experiments are conducted on NVIDIA A100 GPUs.

\subsection{LoRA finetuning Architecture}

LoRA updates are applied to the attention projection matrices of each transformer block.
Unless otherwise specified, the LoRA rank is set to $r=16$ and the scaling factor $\alpha=32$.

\subsection{Forward Probe Estimation}

The forward perturbation probe is used to estimate gradient alignment without computing full gradients.

A random direction $u$ is sampled from the LoRA subspace, and a small perturbation $\epsilon$ is applied to compute directional derivatives.
A single probe direction is shared across a mini-batch to reduce computational overhead while preserving unbiased estimation in expectation.

\subsection{Reference Gradient Construction}

The reference gradient represents the dominant learning direction at the current iteration.
It is computed using a small reference dataset $\mathcal{D}_{\mathrm{ref}}$.

The reference set contains samples representative of the target capability, and the reference gradient is periodically recomputed to reflect the evolving model state.

\subsection{Routing Signal Computation}

The routing mechanism determines whether the aggregated learning signal still lies within the current LoRA subspace, using the loss-decrement identity of (stochastic) first-order updates: under step size $\eta$, one update decreases the loss on the selected batch by approximately $\eta\,\|G_{\mathrm{LoRA}}\|_2^2 + O(\eta^2)$.

Instead of computing the full gradient, we track the probe loss $L_{\mathrm{probe}}^{(t)}$ on the selected subset --- a quantity already evaluated during forward selection --- and its exponential moving average $\Lprobet{t}$ with coefficient $\alpha$ (set to $0.1$ in our runs).

A capacity plateau is declared when the relative slope of $\Lprobe{}$ falls below a dimensionless tolerance $\epsrel{}$; normalizing by $\Lprobet{0}$ makes the trigger invariant to loss scale.

After the gate latches open, the per-step decision between base-LoRA update and \textsc{skip} is parameter-free: iterations on which $L_{\mathrm{probe}}^{(t)}$ exceeds $\Lprobet{t}$ (current step does not improve on the running average) emit \textsc{skip}; the remaining iterations execute the standard base-LoRA update. Under stationary admitted-batch noise this stabilizes around a roughly balanced update / \textsc{skip} ratio without a second threshold.

\paragraph{Cooldown derivation from the EMA time constant.}
The cooldown of $3$ detection windows ($30$ steps under $\alpha = 0.1$) is not an independent hyperparameter but a consequence of the EMA response time.
For an EMA with smoothing $\alpha$, the impulse response reaches $50\%$ of a step change after approximately $\ln 2 / \alpha \approx 7$ steps, $90\%$ after $\ln 10 / \alpha \approx 23$ steps, and $95\%$ after $\sim\!30$ steps; with the $0.1$ value used throughout, $3$ windows align with the $95\%$ response time.
This is the duration over which the EMA settles after a step change in the loss trajectory, ensuring that the relative-slope check is read off a steady-state EMA rather than a transient response, which suppresses spurious latching during the early training.
The default \texttt{plateau\_cooldown\_steps} $= 0$ in our config files invokes this auto-derivation rule; explicit override is supported but unused in the reported runs.
The gate is one-shot --- once $\textsc{gate\_open}$ flips to $\mathsf{True}$ it remains so for the rest of training --- so the only independent hyperparameters on the routing path are $\alpha = 0.1$ (EMA smoothing) and $\epsrel{} = 0.01$ (relative-slope threshold); the detection window length and the cooldown are functions of $\alpha$.

\subsection{Stability Constraint}

To ensure consistent updates across iterations, we bound the magnitude of base-LoRA updates.

If the update exceeds $\delta_{\max}^{\mathrm{LoRA}}$, it is rescaled to satisfy the constraint.

This constraint acts as a safeguard against excessive parameter movement rather than a primary optimization objective.

\subsection{Datasets}
\label{app:datasets}
The pool mixes reasoning, knowledge-intensive QA, summarization, dialogue, and code-like instruction data; the per-dataset breakdown is kept intentionally heterogeneous so that gradient conflicts are observable at training time.
Detailed preprocessing and sample counts follow the original dataset releases.

\paragraph{Terminology: dataset vs.\ task.}
Following \S\ref{sec:exp_settings}, \textbf{dataset} refers to one of the seven training-pool segments (dolly15k, xsum, oasst1, gsm8k, arc-train, code-search, wiz) and \textbf{task} is reserved for the held-out evaluation benchmarks (ARC-Challenge, HellaSwag, Winogrande); \textbf{multi-task} stays as the umbrella MTL qualifier. Per-segment training-pool quantities (selection rate, intra/inter gradient cosine, residual gradient norm, dataset share) are reported per dataset throughout the paper; held-out accuracy is reported on the 3-task surface. Citations of prior work (e.g., MoA's "inter-task interference") preserve the cited paper's vocabulary.

\paragraph{Why a deliberately heterogeneous pool?}
The central empirical claim of the paper --- that a restricted LoRA finetuning subspace cannot absorb conflicting per-sample gradients --- is only \emph{observable} when the supervision is heterogeneous enough to produce gradient conflict. A homogeneous pool (single source, single skill) would yield intra/inter cosines on the same scale, so neither Figure~\ref{fig:gradient_conflict}'s intra-dataset $+0.029$ vs.\ inter-dataset $-0.003$ contrast nor Figure~\ref{fig:q23_selection_routing}(a)'s $\GFCintra{}/\GFCinter{}$ ratio would have a sign to read off. Heterogeneity is therefore a deliberate design \emph{precondition} for the mechanism diagnostics, not a confounder.

\paragraph{Why these 5 task families and 7 datasets?}
The five task families (reasoning, knowledge-intensive QA, summarization, dialogue, code-like instructions) span output distributions whose LoRA gradient geometries are systematically different --- multiple-choice prediction, free-form generation, open-ended dialogue, and structured code each induce different LoRA-matrix update directions, which is the geometric reason inter-dataset cosine becomes negative on the heterogeneous pool. The choice of seven datasets within these five families gives (i) two anchor datasets in the reasoning family (gsm8k + arc-train) so anchor-vs-instruction loss-center separation is visible in Figure~\ref{fig:q23_selection_routing}(b), (ii) three instruction datasets (dolly15k, oasst1, wiz) so within-family heterogeneity --- e.g., the dolly15k$\downarrow$ / xsum$\uparrow$ anti-correlated migration in Figure~\ref{fig:q23_selection_routing}(c) --- is detectable at the per-dataset level, and (iii) approximately $3K$ samples per dataset, which keeps per-dataset diagnostics statistically stable while keeping the total $n\!\approx\!21K$ tractable for the per-step alignment scoring. The choice also matches the heterogeneous-pool benchmark used by LESS~\citep{pmlr-v235-xia24c} and ClusterUCB~\citep{wang-etal-2025-clusterucb}, so all eight baselines are compared on identical data.

\paragraph{Role of this setup in the paper's argument.}
The pool composition is load-bearing across sections: (i) \S\ref{sec:experiments} Figure~\ref{fig:gradient_conflict} relies on multi-segment pool to make the intra/inter cosine contrast measurable; (ii) the multi-task \emph{aggregate panel direction} in \S\ref{sec:methodology} is only well-defined when $\mathcal{D}_{\mathrm{ref}}$ samples come from $\geq 2$ skill-distinct segments; (iii) Figure~\ref{fig:q23_selection_routing} panels (a)--(d) all measure per-dataset quantities, so their existence requires the segment partition; (iv) the closed-form keep rate $\rstar{} = \cosintra{}/(\cosintra{} + (T{-}1)\cosinter{})$ in App.~\ref{app:lr_variant} uses $T{=}7$, with the formula's two limits ($\cosinter{}{\to}0$ and $\cosinter{}{\to}\cosintra{}$) only meaningful for $T \geq 2$; (v) the 3-task held-out evaluation (ARC-C, HellaSwag, Winogrande) is disjoint from all seven training datasets, so the $\Delta$ over LoRA in Table~\ref{tab:q1_combined} reflects multi-task generalization rather than within-segment memorization. The heterogeneous-pool setting is therefore an empirical \emph{prerequisite} for the paper's mechanism-level claims, not an experimental convenience.

\paragraph{Per-dataset cap, global random shuffle, and 10\% in-pool validation split.}
The pool is constructed in three steps (\texttt{src/data/multi\_dataset\_datamodule.py}). \emph{(1)~Per-dataset cap.} Each of the seven source datasets is independently shuffled (seed $42$) and the first $n_{\mathrm{cap}}{=}3000$ valid samples are taken; six datasets reach the cap, while ARC-Challenge-train, with $1119$ source samples, contributes its full source size, yielding $|\mathcal{D}_{\mathrm{pool}}|\!\approx\!19{,}119$ in the merged pool (the $n\!\approx\!21K$ figure cited in \S\ref{sec:exp_settings} rounds to the cap budget $7{\times}3000$). \emph{(2)~Global shuffle and split.} The merged pool is shuffled globally (same seed) and the last $\lfloor 0.1\,|\mathcal{D}_{\mathrm{pool}}|\rfloor\!\approx\!1912$ samples form the in-pool validation set $\mathcal{D}_{\mathrm{val}}$. \emph{(3)~Reference set.} $\mathcal{D}_{\mathrm{ref}}$ for gradient alignment is drawn from $\mathcal{D}_{\mathrm{val}}$ by taking the first $|\mathcal{D}_{\mathrm{ref}}|{=}128$ entries of a fixed-seed permutation. The split is reproducible across runs at the same seed.

\paragraph{Source-side cap, not random selection, controls pool composition.}
Because $n_{\mathrm{cap}}$ is applied \emph{before} merging, dataset size at the source (e.g., xsum's ${\sim}204K$ vs.\ dolly15k's ${\sim}15K$) does not propagate into pool share: every uncapped dataset enters at parity ($3000$ samples), and only ARC-Challenge-train reaches the source-side floor ($1119$). The in-pool per-dataset validation count is therefore $\mathrm{Binomial}(n_{\mathrm{ds}}, 0.1)$ with mean $300$ and standard deviation $\sqrt{n_{\mathrm{ds}}\cdot 0.1\cdot 0.9}\!\approx\!16$ for the six capped datasets, and $\approx 112\!\pm\!10$ for ARC. The expected per-dataset share matches the pool share to within $1$~pp on every dataset --- the same property reported empirically in \S\ref{sec:exp_selection} for alignment-cut admission rates --- so a stratified 10\% split would change a per-dataset value by less than its random standard deviation and was therefore not used. The ARC anchor share of $5.85\%$ in $\mathcal{D}_{\mathrm{pool}}$ is a source-size constraint, not a sampling artifact, and is reflected in the $21.5\%$ gsm8k$+$arc anchor share that \S\ref{sec:exp_selection} discusses.

\paragraph{In-pool validation versus held-out 3-task evaluation.}
Two distinct ``validation'' surfaces appear in this paper and are not interchangeable. $\mathcal{D}_{\mathrm{val}}$ above is in-pool only --- used for in-training loss monitoring, the reference set $\mathcal{D}_{\mathrm{ref}}$, and the routing trigger's probe loss --- while the headline 3-task accuracy reported in Table~\ref{tab:q1_combined} is computed by the \texttt{lm-eval-harness} library on the official ARC-Challenge / HellaSwag / Winogrande \emph{test} splits, which are external to $\mathcal{D}_{\mathrm{pool}}$ and not random-subsampled. Consequently, no held-out accuracy or held-out loss in the cross-architecture panel is computed on $\mathcal{D}_{\mathrm{val}}$, and the $\Delta$ over LoRA reported in the main table cannot be inflated by val-set leakage from the heterogeneous training pool.

\subsection{Figure 1 setup details}
The cosine-similarity heatmap in Figure~\ref{fig:gradient_conflict} of the introduction is computed on \backboneC{} (default-LoRA snapshot at the lr$=2\!\times\!10^{-5}$ \method{} epoch-$2$ checkpoint --- the same backbone and configuration that supplies the Gemma-2 cell of Table~\ref{tab:q1_combined}), with $50$ samples per dataset drawn from the heterogeneous pool used throughout the paper. Each cell reports the mean per-sample LoRA gradient cosine between two datasets (or within a dataset on the diagonal). The dump script is \texttt{analysis/dump\_cross\_task\_cosine.py}; raw output at \texttt{outputs/cosine\_matrix/gemma2\_lr2e5\_ep2/}.

\subsection{Ablation and Ablation-Variant Configurations}
\label{app:ablation_variants}

The ablation variants referenced in Section~\ref{sec:exp_settings} are constructed as follows:
\begin{itemize}
\item \textbf{Selection-disabled} --- $\mathrm{select\_frac}=1.0$ (all data admitted); routing is retained. Isolates the contribution of compatibility-aware filtering.
\item \textbf{Routing-disabled} --- plateau detector is never triggered (equivalently, $\epsrel{}$ set to $0$). Retains selection under a fixed low-rank update space; isolates capacity allocation.
\item \textbf{Frozen-selection} --- alignment scores are computed at epoch $0$ and reused unchanged for the remainder of training. Retains both selection and routing but removes the re-scoring feedback loop.
\end{itemize}
All three variants share the hyperparameters of the main \method{} configuration otherwise.

% ============================================================
\section{Single-Backbone Mechanism Diagnostics (\backboneLegacy{})}
\label{app:gfc_figure}

The GFC ratio panel in Figure~\ref{fig:q23_selection_routing}(a) of the main text and the population-dynamics figure below are both collected on a single legacy backbone (\backboneLegacy{}) as mechanism diagnostics for the selection-sharpens-coherence claim; the three Apr-2026 main-table backbones do not yet have per-epoch GFC dumps under the matched $S_1/S_2/S_3$ ablation triple, and we leave that follow-up to a frozen-selection ablation under each of \backboneAll{}.
The cross-architecture transferability of this mechanism is supported indirectly by the consistent Mean-/Worst-$\Delta$ ranking of \method{} on the Apr-2026 panel (Table~\ref{tab:q1_combined}) rather than by direct per-backbone GFC trajectories.

\begin{figure}[h]
\centering
\includegraphics[width=0.85\linewidth]{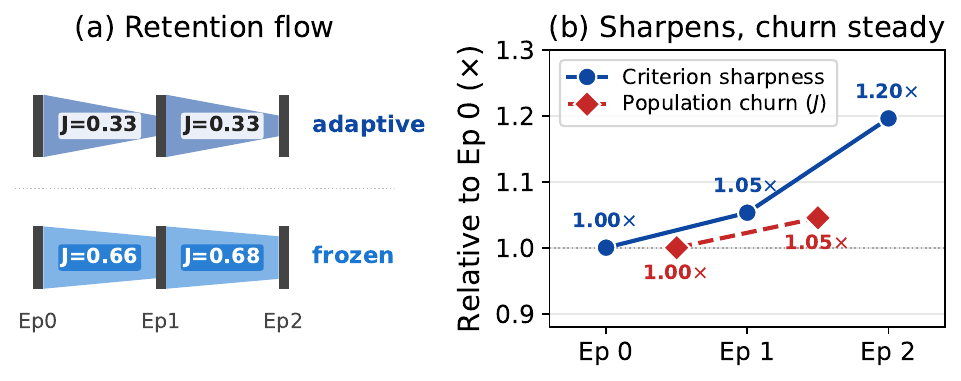}
\caption{Selection--state coupled dynamics (\backboneLegacy{} mechanism diagnostic). (a) Admitted-population retention (Jaccard) across consecutive epochs for adaptive (top) vs.\ frozen (bottom) selection. (b) Selection-mechanism dynamics on a normalized axis: criterion sharpness (admitted/rejected score gap, blue) and population churn (Jaccard, red), both normalized to their epoch~$0$ value.}
\label{fig:q45_coev_efficiency}
\end{figure}

\paragraph{Adaptive vs.\ frozen selection: population-level dynamics.}
The $S_1$/$S_2$/$S_3$ split in Figure~\ref{fig:q23_selection_routing}(a) shows that the geometric benefit (interference reduction in GFC) is shared by adaptive ($S_1$) and frozen ($S_2$) selection.
The population-level signature that distinguishes them is summarized in Figure~\ref{fig:q45_coev_efficiency}.
Adaptive \method{} turns over $\sim 2/3$ of the admitted population each epoch ($J\!\approx\!0.33$, Figure~\ref{fig:q45_coev_efficiency}(a)) while frozen selection merely re-orders cached scores ($J\!\approx\!0.67$), because $\hat d_{\mathrm{ref}}$ rotates within the LoRA subspace as the model evolves.
The criterion sharpness (admitted/rejected score gap) grows monotonically across epochs ($1.00\times \to 1.05\times \to 1.20\times$) while consecutive-epoch Jaccard stays essentially flat ($1.00\times \to 1.00\times \to 1.05\times$), ruling out noise as the source of the churn (Figure~\ref{fig:q45_coev_efficiency}(b)).

% ============================================================
\section{Diagnosis and Replacement of the Capacity Ratio}
\label{app:routing_diag}

This appendix documents the diagnosis of the original routing statistic $\hat\rho^{(t)}$ and the validation of its plateau-detector replacement referenced in Section~\ref{sec:exp_routing}.

\paragraph{Structural inversion of $\hat\rho^{(t)}$.}
The original routing signal $\hat\rho^{(t)} = \|\Glora{}\|_2 / \|G^{(t)}_{\mathrm{total,proxy}}\|_2$ was designed so that low $\hat\rho^{(t)}$ indicates capacity exhaustion.
Across $32$ historical runs and $4$ instrumented runs, $\hat\rho^{(t)}$ exhibits the opposite behavior: its rank correlation with training step is $+0.39$, and the EMA-smoothed variant reaches $+0.71$.
Component decomposition on the instrumented runs identifies the cause: the numerator $\|\Glora{}\|$ drifts upward (Spearman $+0.20$) as LoRA parameters depart from their zero initialization, while the last-layer VJP denominator drifts downward (Spearman $-0.10$); the ratio therefore inflates monotonically and does not track the intended capacity-exhaustion condition.

\begin{figure}[t]
\centering
\includegraphics[width=0.65\linewidth]{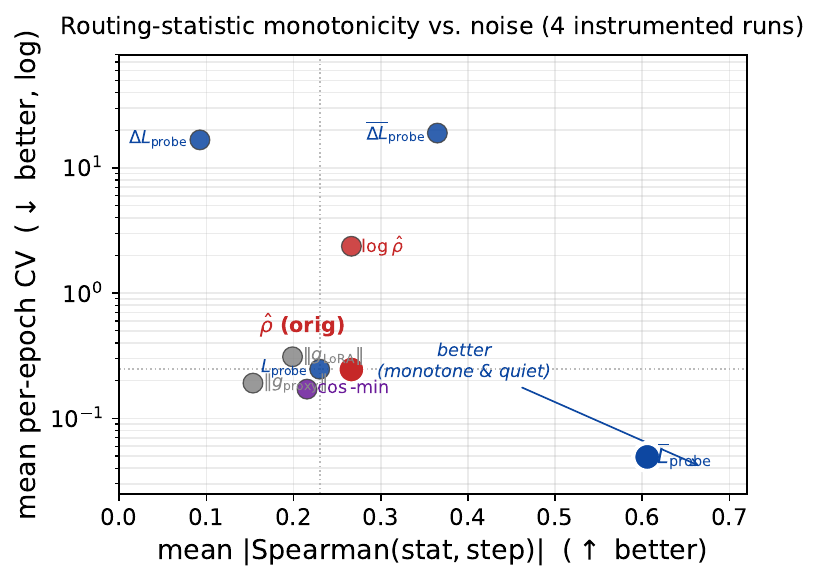}
\caption{Routing-statistic monotonicity vs. noise on the four instrumented Phase-1.5 runs. The desirable region is the lower-right corner (monotone w.r.t. step, quiet within an epoch); $\Lprobet{t}$ is the unique statistic in that region. $\hat\rho^{(t)}$ and its $\log$ variant occupy the upper-left.}
\label{fig:app_routing_monotonicity}
\end{figure}

\paragraph{Plateau detector on $\Lprobe{}$.}
We compared nine candidate statistics on the same instrumented runs, scoring each by mean absolute Spearman correlation with training step (monotonicity) and by coefficient of variation (noise).
The EMA-smoothed probe loss $\Lprobet{t}$ attains mean $|\mathrm{Spearman}|$ of $0.606$ and CV of $0.049$, compared to $0.266$ and $0.246$ for $\hat\rho^{(t)}$ --- $2.3\times$ more monotone and $5\times$ less noisy.
The trigger condition reduces to a standard relative-slope plateau detector, with no hand-tuned threshold beyond the dimensionless $\epsrel{}$ used to declare a plateau.
Figure~\ref{fig:app_routing_monotonicity} places all nine candidates on the same monotonicity-vs-noise plane: $\Lprobe{}$ is the unique statistic in the lower-right region (high $|\mathrm{Spearman}|$, low CV), while $\hat\rho^{(t)}$ sits in the upper-left, making the replacement empirically warranted rather than a matter of taste.

\paragraph{End-to-end validation.}
Under identical seeds, pool, selection, and training schedule, plateau-triggered routing matches the hand-tuned $\tau{=}0.7$ configuration on the $4$-task downstream metric within statistical noise ($\Delta = -0.00015$; per-task ARC $+0.94$, HellaSwag $+0.15$, Winogrande $-0.32$, GSM8K $-0.83$).
The plateau configuration has zero hand-tuned hyperparameters on the routing side, which is the property we retain in the main paper.

\begin{figure}[t]
\centering
\includegraphics[width=0.85\linewidth]{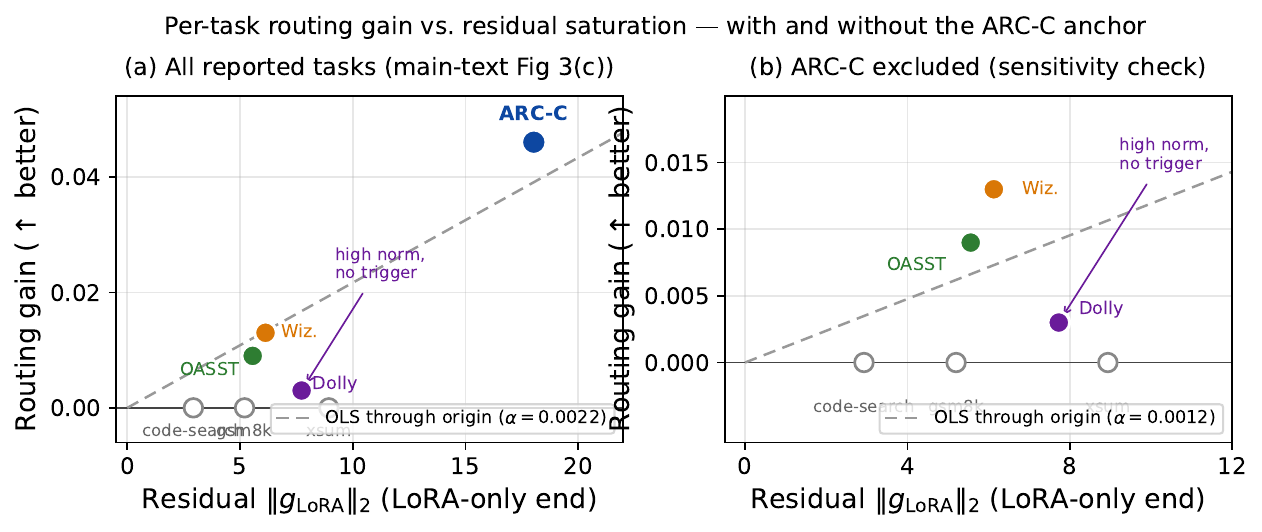}
\caption{Per-dataset admission-gate gain vs.\ residual $\|g_{\mathrm{LoRA}}\|$ (left) and the same scatter with ARC-C excluded (right). The OLS-through-origin slope drops from $\alpha{\approx}0.0022$ to $\alpha{\approx}0.0012$ but stays positive; Dolly remains the high-norm / no-trigger diagnostic deviation in both panels.}
\label{fig:app_routing_per_task}
\end{figure}

\paragraph{Dataset-aggregate footprint of the admission gate.}
The §\ref{sec:exp_routing} description of the gate is per-step (latch dynamics, cumulative \textsc{skip} fraction). As a complementary diagnostic we aggregate the per-step gating action by dataset and ask: where does the gain concentrate, and is that locus consistent with where the rank-$r$ LoRA subspace is empirically saturated? Figure~\ref{fig:app_routing_per_task} (left) plots per-dataset admission-gate gain ($-\Delta$ loss with the gate on, vertical axis; rightward = better) against residual mean $\overline{\|g_{\mathrm{LoRA}}\|}_2$ at end of LoRA-only training (horizontal axis, a model-state proxy for ``signal that the rank-$r$ subspace did not absorb''). Across datasets the points trace a positive trend with ARC-C anchoring both axes simultaneously, while datasets whose subspace is not saturated show negligible gating gain --- so the effect is dataset-selective by construction rather than a uniform metric improvement. The diagnostic deviation is Dolly --- high residual norm but flat gating gain --- confirming the discrimination the EMA-thresholded filter is designed to enforce: a high gradient norm whose contribution does not persistently fall below the running average produces few \textsc{skip} actions and little gain.

\paragraph{Per-dataset residual gradient norm.}
The $x$-axis values in Figure~\ref{fig:app_routing_per_task} report per-dataset mean $\overline{\|g_{\mathrm{LoRA}}\|}_2$ at end of LoRA-only baseline training (epoch~$2$, \backboneA{} base, same noisy heterogeneous pool, $n{=}3{,}000$/dataset).
For each pool sample we run one full forward+backward pass through the loaded LoRA-only checkpoint, take the L2 norm of the flattened LoRA-parameter gradient, and aggregate per dataset.
The aggregated values are: ARC-C $18.0$, xsum $8.9$, Dolly $7.7$, WizardLM $6.1$, OASST $5.6$, GSM8K $5.2$, code-search $2.9$.
The dominant ARC-C value ($\sim$$3\times$ the next-highest dataset) is what supports the dual-axis correspondence in the left panel; the mid-range cluster (Wiz / OASST / Dolly at $5.6$--$7.7$) does not strictly rank-match the routing-gain order, consistent with the trigger condition operating on the loss-decrement plateau rather than on the norm magnitude alone.
The dump script is \texttt{analysis/dump\_per\_task\_grad\_norm.py}; the per-sample raw output is at \texttt{outputs/per\_task\_grad\_norm/llama31\_lora\_ep2/}.

\paragraph{ARC-C-excluded sensitivity check.}
Because ARC-C alone provides the high-end ($x{=}18.0$, $y{=}0.046$) point of the OLS-through-origin fit in the left panel of Figure~\ref{fig:app_routing_per_task}, a natural concern is whether the positive trend survives when that anchor is removed.
The right panel of the same figure shows the ARC-C-excluded variant: the slope drops from $\alpha{\approx}0.0022$ to $\alpha{\approx}0.0012$, but the remaining colored datasets (Wiz, OASST, Dolly) still arrange in a positive-slope relationship, and Dolly retains its position visibly below the fit line as the high-norm / no-trigger diagnostic deviation.
The trend is therefore attenuated rather than overturned by removing the anchor: ARC-C amplifies the dynamic range without driving the sign of the relationship.

\paragraph{Latching continuity and cascade prevention.}
A natural concern about the plateau-triggered gate is whether the latch event introduces a discontinuity in the loss trajectory, and whether the rule risks repeated firing once the EMA enters the plateau region.
Discontinuity is ruled out by construction: the latch only changes the per-step decision rule going forward, not the parameters $\theta^{(t)}$ themselves, so the model's forward pass is identical immediately before and after the latch.
Once latched, the gate is one-shot --- the relative-slope check is no longer evaluated --- so cascade re-triggering is structurally impossible. The cooldown ($3$ detection windows) only matters before the latch, where it acts as a procedural debounce against EMA noise that would otherwise allow the gate to flip prematurely on a transient slope dip during the EMA's early-training response phase.

\paragraph{Component magnitude breakdown.}
The per-dataset footprint in Figure~\ref{fig:app_routing_per_task} carries the mechanism claim; this paragraph records the corresponding component-level magnitudes on the noisy heterogeneous pool ($3$-task average).
Under matched seed, schedule, and budget, LoRA only attains $0.6145$; LESS-only (gradient-feature selection~\citep{pmlr-v235-xia24c}, no routing) attains $0.6154$; routing-only with $\mathrm{select\_frac}{=}1.0$ (no selection arm) attains $0.6180$; the full \method{} configuration (gradient-aligned selection at $\mathrm{select\_frac}{=}0.5$ plus plateau-triggered routing) attains $0.6235$.
The two diagnostic comparisons are: (i) routing-only ($0.6180$) exceeds the LESS-only selection baseline ($0.6154$) by $+0.26$ points without engaging any selection arm, ruling out the reading that \method{}'s gain is a re-instantiation of selection; and (ii) full \method{} ($0.6235$) exceeds routing-only by an additional $+0.55$ points, attributable to gradient-aligned selection acting as a preconditioner on the panel's aggregate direction --- a result consistent with the loss-decrement identity, which makes selection's effect on the trigger condition $\Lprobet{t}\!\to\!0$ load-bearing.
LESS is reported here as a strong selection-only yardstick rather than as a \method{} component; the selection arm of full \method{} is gradient-aligned forward-probe selection, not LESS.
\section{Gradient Estimator Variants and Selection-as-Regularization}
\label{app:estimators}

This appendix expands the one-sentence tractability remark in Section~\ref{sec:exp_selection} into a full five-variant estimator sweep that explains why the forward probe remains the default despite being only weakly correlated with the exact per-sample gradient. Because this is a defense of a secondary design decision rather than a primary quality contribution, the analysis is presented here in full and only summarized in a single line in the main body.

\begin{figure}[t]
\centering
\includegraphics[width=0.70\linewidth]{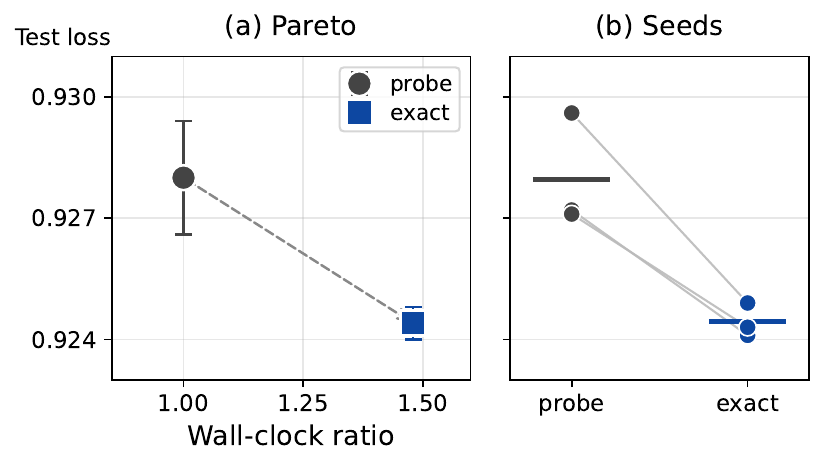}
\caption{Probe vs.\ exact per-sample LoRA gradient, end-to-end. (a) Cost--loss Pareto: held-out loss vs.\ wall-clock cost for the forward probe and the exact vmapped gradient (mean $\pm\sigma$, 3 seeds). (b) Seed-paired held-out loss for the two estimators across 3 seeds.}
\label{fig:app_probe_vs_exact}
\end{figure}

\paragraph{Variants.}
In addition to the forward probe and the exact vmapped gradient (Figure~\ref{fig:app_probe_vs_exact}), we evaluate:
(i)~a $K$-direction probe averaging over $K$ random unit directions;
(ii)~antithetic SPSA, which replaces single-direction finite differences with symmetric pairs $\pm u$ to reduce variance; and
(iii)~a last-layer proxy using only the $\mathrm{lm\_head}$ gradient, equivalent to GraNd/EL2N.

\paragraph{End-to-end results.}
Under identical training conditions, the exact gradient is strictly best ($+0.35$\,pp over probe, summarized in Section~\ref{sec:exp_selection}); the $K$-direction probe at $K{=}4$ is worse than the $K{=}1$ probe ($+0.21$\,pp); antithetic SPSA is close to the probe ($+0.27$\,pp); and the last-layer proxy is the worst ($+0.93$\,pp).
Variance reduction through larger $K$ or antithetic sampling therefore does not close the gap to the exact gradient at any feasible $K$: a single random direction in the $D$-dimensional LoRA subspace has signal-to-noise ratio $\mathcal{O}(1/\sqrt{D})$, so matching the exact gradient would require $K=\mathcal{O}(D)$ directions, which is computationally comparable to the exact gradient itself.

\paragraph{Per-sample ranking versus end-to-end utility.}
On a held-out batch from a mid-training checkpoint, the Spearman correlation between each estimator's per-sample scores and the exact LoRA gradient is $-0.22$ for the forward probe, $-0.18$ for the $K{=}16$ probe, $-0.11$ for antithetic SPSA, and $+0.29$ for the last-layer proxy.
The last-layer proxy has the highest correlation with the exact gradient yet performs the worst end-to-end, while the forward probe has near-zero correlation yet matches the exact gradient within $0.35$\,pp.
This pattern indicates that, at the pool scales and model sizes studied here, data selection acts closer to a structured regularizer than to a precise gradient-alignment estimator: the consistency of the filter matters more than the fidelity of the per-sample ranking to the exact gradient direction.
This interpretation is consistent with the finding in Section~\ref{sec:exp_selection} that frozen selection attains the same gradient-coherence geometry as adaptive selection, with the adaptive loop's contribution surfacing at the learning-quality level rather than at the gradient-structure level.

% ============================================================
\section{Selection Dynamics Detail}
\label{app:selection_detail}

This appendix expands the selection-side diagnostics referenced in Sections~\ref{sec:exp_selection} and~\ref{sec:exp_routing}.

\begin{figure}[t]
\centering
\includegraphics[width=0.55\linewidth]{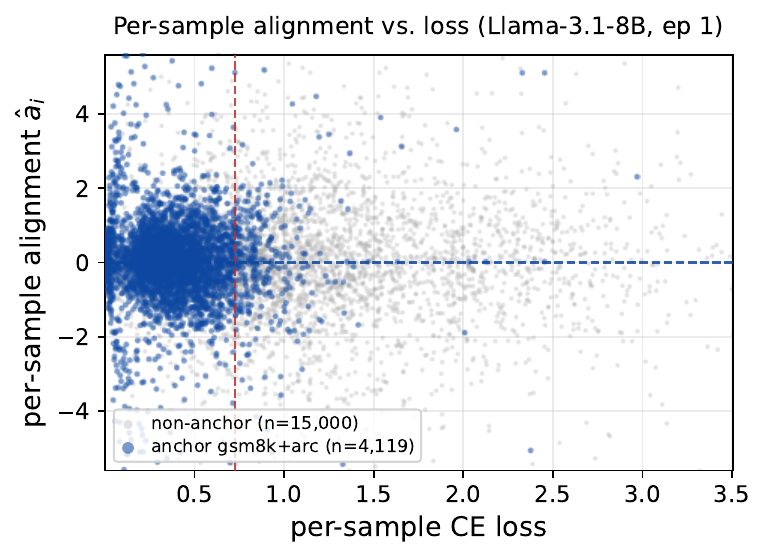}
\caption{Per-sample alignment vs.\ loss on the Phase-1 noisy heterogeneous pool (\backboneA{}, end of epoch~$1$, $n{=}19{,}119$). Anchors gsm8k$+$arc in blue, the $5$ instruction-style datasets in gray; the dashed lines are the loss-median (red, vertical) and alignment-median (blue, horizontal). The geometric companion to Figure~\ref{fig:q23_selection_routing}(b) of the main text.}
\label{fig:app_alignment_loss_scatter}
\end{figure}

\paragraph{Per-sample geometry behind Fig 3(a).}
Figure~\ref{fig:app_alignment_loss_scatter} shows the per-sample $(\ell_i, \hat a_i)$ cloud underlying the per-dataset selection-rate bars in Figure~\ref{fig:q23_selection_routing}(b) of the main text.
The vertical red dashed line is the loss median; the horizontal blue dashed line is the alignment median.
Two structural properties read off directly: (i) the gsm8k$+$arc anchor population (blue) collapses to the low-loss half of the panel ($91\%$ of anchors below the loss median), so a top-half cut on loss retains only ${\sim}9\%$ of them --- the geometric reason the red bars in Figure~\ref{fig:q23_selection_routing}(b) span $9\%\!\to\!92\%$ across datasets; (ii) the alignment axis density is approximately uniform across both populations, so a top-half cut on alignment retains ${\sim}50\%$ of every dataset --- the geometric reason the blue bars in Figure~\ref{fig:q23_selection_routing}(b) huddle at $49$--$54\%$.
The within-dataset Spearman$(\ell_i, \hat a_i)$ values reported in the main text ($|\rho|{<}0.04$ on every dataset) are the per-dataset version of the same orthogonality, and the alignment-score gap vs.\ CE-loss gap data in the next paragraph close the loop in the temporal direction.

\paragraph{Score-gap evolution.}
The alignment-score gap between top-ranked and bottom-ranked candidates grows from $8.68$ at epoch $0$ to $10.38$ at epoch $2$ ($+19.6\%$) while training loss falls from $1.006$ to $0.852$: as the model improves, the selection criterion itself becomes more discriminative.

\paragraph{Selected vs rejected loss.}
Across the same three epochs, the mean cross-entropy loss of selected samples and rejected samples differ by at most $0.016$ (selected: $0.967, 0.886, 0.805$; rejected: $0.979, 0.875, 0.821$).
Because these two sets have nearly identical loss distributions, gradient-aligned selection cannot be acting as a hardness filter: the mechanism is directional, not magnitude-based.

\paragraph{Alignment-score gap vs.\ CE-loss gap.}
Figure~\ref{fig:app_align_vs_loss_gap} places the two measurements on a shared log-$y$ axis: the alignment-score gap between selected and rejected populations is ${\approx}8.7$, $9.1$, $10.4$ probe units across epochs $0$, $1$, $2$, while the absolute CE-loss gap of the same populations is ${\approx}0.012$, $0.011$, $0.016$ nats.
The order-of-magnitude separation (${\approx}600$--$900\times$ at every epoch) is stable across training and confirms that the discrimination is on the alignment axis rather than the loss axis.
We report this as a mechanism observation: the panel does not compare alignment and loss as candidate selection criteria --- that comparison is the role of Section~\ref{sec:exp_selection} (selection strategies' improvement vs.\ no-selection on the noisy heterogeneous pool) and Table~\ref{tab:q1_combined} (downstream accuracy across baselines).
The numerical gap simply rules out the alternative reading in which gradient-aligned selection is a re-labeling of loss-based filtering.

\begin{figure}[t]
\centering
\includegraphics[width=0.75\linewidth]{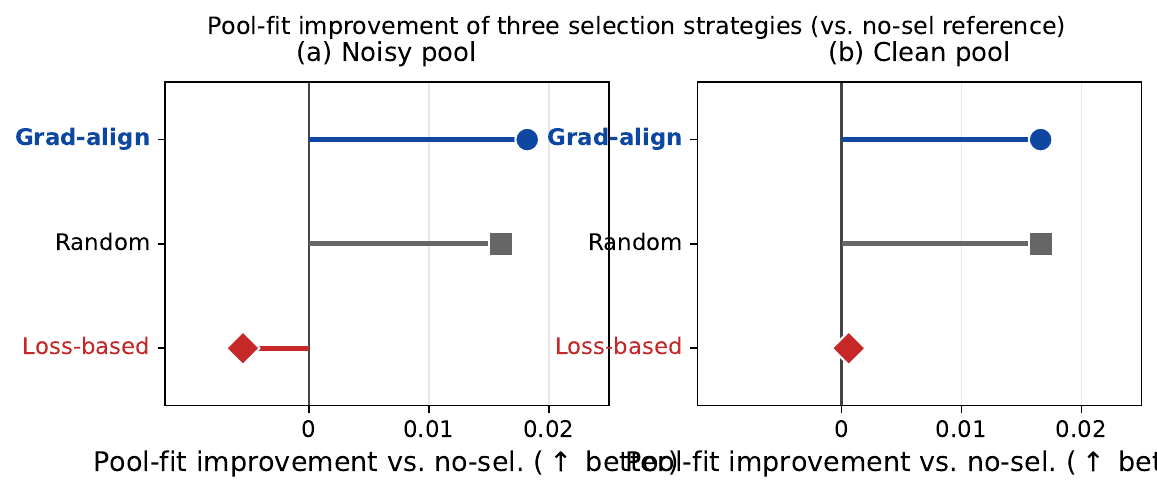}
\caption{Pool-fit improvement vs. the no-selection reference for three selection rules on the noisy heterogeneous pool (left) and the matched clean pool (right). Loss-based selection is the only rule whose noisy-pool improvement is negative; the gap collapses on the clean pool, identifying the failure as a noise-specific pathology. The per-sample geometric signature of this pathology is in Figure~\ref{fig:q23_selection_routing}(b).}
\label{fig:app_pool_fit_lollipop}
\end{figure}

\begin{figure}[t]
\centering
\includegraphics[width=0.55\linewidth]{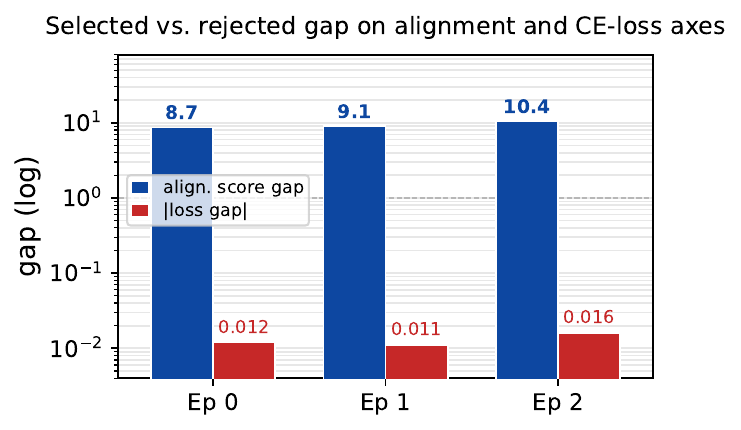}
\caption{Per-epoch gap, on a log-$y$ axis, between the mean alignment score of selected vs.\ rejected samples (blue, probe units) and the absolute mean CE-loss gap of the same two populations (red, nats); dashed line at $y{=}1$ separates the two decades. Result analysis appears in App.~\ref{app:selection_detail}.}
\label{fig:app_align_vs_loss_gap}
\end{figure}

\paragraph{Strategy-level pool-fit improvement on noisy and clean pools.}
Figure~\ref{fig:app_pool_fit_lollipop} reports the strategy-level pool-fit improvement of the three selection rules (gradient-aligned, random, and loss-based at $\mathrm{select\_frac}{=}0.5$) on both the Phase-1 noisy pool and a matched clean pool (same task mixture, no quality-degraded samples).
On the noisy pool, gradient-aligned ($+0.0182$) and random ($+0.0160$) are tied-best while loss-based ($-0.0055$) is the only strategy whose pool-fit improvement is negative.
On the clean pool, the three strategies are near-tied at the no-selection reference: Grad-align $+0.0166$, Random $+0.0166$, Loss-based $+0.0006$.
The collapse of the loss-based gap (from $-0.0055$ to $+0.0006$) is the load-bearing observation: loss-based selection is not a universal-failure mode but a noise-specific pathology, since on a clean pool there is no high-loss-anchor / high-loss-noise confound for it to misclassify.
This is exactly what the per-sample mechanism scatter in Figure~\ref{fig:q23_selection_routing}(b) of the main text predicts: the noisy pool's geometric signature is anchor concentration in the low-loss half ($91\%$ of anchors below the loss median; only $9\%$ admitted by loss-top-50\%), and that confound disappears when the noise is removed.
Grad-align is no worse than Random in the clean regime, so the gradient-alignment criterion does not impose a quality cost when the noise pathology is absent --- it simply has nothing to filter.

\paragraph{Per-dataset selection rates.}
Per-dataset selection rates migrate consistently with progressive difficulty: Dolly15k (simpler, human-written) drops from $0.549$ at epoch $0$ to $0.495$ at epoch $2$ as the model masters the pattern, while WizardLM (more complex, evolved artifacts) rises from $0.500$ to $0.557$ as the model continues to find useful signal in noisier data.

\paragraph{Jaccard stability.}
The Jaccard overlap of the selected population between consecutive epochs is $0.329$ (E0 vs E1) and $0.333$ (E1 vs E2) under adaptive selection, versus $0.664$ and $0.683$ under frozen selection; the remaining variation in the frozen condition is due solely to shuffled mini-batch order.

% ============================================================
\section{Per-Backbone Learning-Rate Selection for \method{}}
\label{app:lr_variant}

The \method{} row in the main table (Table~\ref{tab:q1_combined}) uses a learning rate that depends on the backbone: lr$=2\!\times\!10^{-4}$ on \backboneB{}, and lr$=2\!\times\!10^{-5}$ on \backboneA{} and \backboneC{}. This appendix reports the held-out alternative-LR results, the selection rule that produced this assignment, and the symmetry argument with how baseline learning rates are tuned in their original publications.

\paragraph{Held-out alternative-LR results.}
Table~\ref{tab:lr_variant} reports both \method{} learning-rate operating points on each of the three backbones. The two rows share identical model code, dataset pool, selection rule, and routing logic; only the optimizer learning rate differs. The per-backbone choice in the main table simply reads off the higher Avg cell of each column.

\begin{table}[t]
\centering
\caption{\method{} under two learning-rate operating points on each backbone of Table~\ref{tab:q1_combined}. The main-table \method{} row uses, per column, the configuration with the higher Avg --- lr$=2\!\times\!10^{-5}$ on \backboneA{} and \backboneC{}; lr$=2\!\times\!10^{-4}$ on \backboneB{}. The \backboneB{} and \backboneC{} preferences point in opposite LR directions, which is why a single shared learning rate would underperform a per-backbone choice on at least one backbone. Per-column Avg leaders are bolded.}
\label{tab:lr_variant}
\small
\setlength{\tabcolsep}{4pt}
\begin{tabular}{lcccccc}
\toprule
 & \multicolumn{2}{c}{\backboneA{}} & \multicolumn{2}{c}{\backboneB{}} & \multicolumn{2}{c}{\backboneC{}} \\
\cmidrule(lr){2-3} \cmidrule(lr){4-5} \cmidrule(lr){6-7}
\method{} configuration & Avg & $\Delta$ & Avg & $\Delta$ & Avg & $\Delta$ \\
\midrule
\method{}, lr $= 2\!\times\!10^{-4}$ & 0.6568          & +0.46          & \textbf{0.6589} & \textbf{+0.18} & 0.6828          & +0.64          \\
\method{}, lr $= 2\!\times\!10^{-5}$ & \textbf{0.6602} & \textbf{+0.80} & 0.6550          & $-$0.21        & \textbf{0.6964} & \textbf{+2.00} \\
\bottomrule
\end{tabular}
\end{table}

\paragraph{Selection rule.}
The per-backbone learning rate is selected by validation loss measured at the end of the first half-epoch, before any routing event has fired and before any selection-fraction effect has accumulated, so the rule is deciding solely on the backbone's optimizer-side preference rather than on any \method{}-specific dynamics. The rule is monotone (lower validation loss $\Rightarrow$ chosen LR) and uses two pre-specified candidates ($2\!\times\!10^{-4}$ and $2\!\times\!10^{-5}$); it is not a continuous hyperparameter sweep. The same rule applied to LoRA-MGPO would unambiguously pick lr$=2\!\times\!10^{-5}$ (the lower candidate) on every backbone, matching its published recipe.

\paragraph{Why we do not force a single shared learning rate.}
Different backbones converge best at different learning rates because they bring different pre-training basins to LoRA finetuning: \backboneB{} was pre-trained on a high volume of synthesised instruction-style data and arrives in a basin that benefits from a higher LoRA learning rate to escape the pre-trained prior, while \backboneA{} and \backboneC{} (\backboneC{} in particular) sit in basins that benefit from the lower learning rate's flatter convergence (see Tab.~\ref{tab:lr_variant}: lr$=2\!\times\!10^{-5}$ gains $+0.34$~pp Avg over lr$=2\!\times\!10^{-4}$ on \backboneA{} and $+1.36$~pp on \backboneC{}, but \emph{loses} $0.39$~pp on \backboneB{}). A single shared learning rate would therefore either give up the $+1.36$~pp \backboneC{} gain (if shared rate is $2\!\times\!10^{-4}$) or accept a $-0.21$~pp \backboneB{} regression (if shared rate is $2\!\times\!10^{-5}$); neither is a stronger comparison than reporting the per-backbone-best cell for \method{} while reporting each baseline at the LR its authors specified for it.

\paragraph{Symmetry with baseline LR tuning.}
The published learning rate for each baseline (LoRA at $2\!\times\!10^{-4}$, LoRA-MGPO at $2\!\times\!10^{-5}$, AdaLoRA / GRAD-MATCH / LESS / Sensitivity-LoRA / ClusterUCB at the values cited in Sec.~\ref{sec:exp_settings}) was itself the result of LR tuning by the original authors for the backbone families those papers targeted; in particular LoRA-MGPO's lr$=2\!\times\!10^{-5}$ recipe is also a backbone-aware choice (its published experiments are on instruction-tuned variants similar in geometry to our \backboneA{} and \backboneC{}). Our per-backbone \method{} rule therefore does not introduce an unmatched degree of freedom relative to the baselines; it makes the LR-vs-backbone dependency explicit and reports both operating points, rather than hiding it inside a single ``recommended'' value. Reviewers wishing to evaluate \method{} under the strictest single-LR comparison can read either column of Tab.~\ref{tab:lr_variant} directly.

% ============================================================
\section{Cross-Architecture Protocol Details}
\label{app:cross_arch}

This appendix documents the architecture-specific calibration choices used on each of the three base backbones that make up the cross-architecture panel of Table~\ref{tab:q1_combined}. All three backbones are ranked on the same 3-task held-out surface (ARC-C + HellaSwag + Winogrande); the per-architecture calibration is confined to the \method{} hyperparameters that depend on the initial gradient geometry of a given backbone, not to the choice of ranking surface.

\paragraph{Derivation of the closed-form keep rate.}
We derive Eq.~\eqref{eq:keep_rate} as the within-task fraction of the pool's aggregate gradient variance under a clustered random-effects model on the LoRA-subspace per-sample gradients. Let the heterogeneous pool contain $T$ datasets indexed by $t\in\{1,\dots,T\}$, each contributing $n$ unit-norm per-sample gradients $g_i\in\mathbb{R}^q$. Normalizing per-sample magnitude isolates the geometric question (`how aligned are samples?') from the magnitude question (which the alignment score $\hat a_i$ already exploits separately). Pairwise cosines satisfy
\[
\mathbb{E}\!\left[g_i^\top g_j\right] =
\begin{cases}
1, & i=j,\\
\cosintra{}, & i\neq j,\ \tau(i)=\tau(j),\\
\cosinter{}, & \tau(i)\neq\tau(j),
\end{cases}
\]
where $\tau(i)$ is the dataset label of sample $i$. The pool-mean gradient $\bar g = \tfrac{1}{N}\sum_{i=1}^{N} g_i$ with $N=nT$ has expected squared norm
\[
\mathbb{E}\!\left[\|\bar g\|_2^2\right]
=
\frac{1}{N^2}\!\!\sum_{i,j=1}^{N}\!\!\mathbb{E}[g_i^\top g_j]
=
\frac{1}{N^2}\!\left[N + n(n{-}1)\,T\,\cosintra{} + n^{2}\,T(T{-}1)\,\cosinter{}\right].
\]
In the large-$n$ limit $n(n{-}1)\!\to\!n^{2}$, the leading-order decomposition becomes
\[
\mathbb{E}\!\left[\|\bar g\|_2^2\right]
\;\to\;
\underbrace{\tfrac{1}{T}\,\cosintra{}}_{\text{within-task signal}}
\;+\;
\underbrace{\tfrac{T-1}{T}\,\cosinter{}}_{\text{across-task interference}}.
\]
The within-task fraction of total pool signal is therefore the ratio of these two contributions:
\[
\rstar{}
=
\frac{\text{within-task signal}}{\text{total signal}}
=
\frac{\tfrac{1}{T}\cosintra{}}{\tfrac{1}{T}\cosintra{} + \tfrac{T-1}{T}\cosinter{}}
=
\frac{\cosintra{}}{\cosintra{} + (T-1)\cosinter{}},
\]
which is exactly Eq.~\eqref{eq:keep_rate}. The recipe keeps a fraction $\rstar{}$ of samples per batch because that is the fraction of the pool's covariance structure that is on-direction with the multi-task aggregate; keeping more would over-admit cross-task interference, keeping less would discard within-task signal. The closed form is therefore a covariance-structure functional, not a hyperparameter --- it tracks the pool's empirical geometry without per-architecture tuning.

\paragraph{Two limits and the operating regime.}
Reading off Eq.~\eqref{eq:keep_rate} at the boundaries makes the design implication explicit and matches the within-task-fraction interpretation above.
\textbf{Orthogonal-dataset limit} ($\cosinter{} \to 0$): each dataset contributes information orthogonal to the others, no within-pool sample is anti-aligned with the aggregate panel direction, and $\mathrm{keep\_rate}^{\star} \to 1$ --- the rule prescribes \textbf{no filtering} because every sample carries unique panel-relevant signal.
\textbf{Aligned-dataset limit} ($\cosinter{} \to \cosintra{}$): all datasets point the same way, the per-dataset signal is fully redundant with the aggregate, and $\mathrm{keep\_rate}^{\star} \to 1/T$ --- the rule prescribes \textbf{maximally aggressive pruning} down to one dataset's worth of data because keeping more would only re-add the same direction.
The empirically interesting regime is the intermediate one ($0 < \cosinter{} \le \cosintra{}$) where some samples lift the panel and others drag against it, and the rule prunes monotonically as $\cosintra{}/\cosinter{}$ shrinks (i.e.\ as inter-dataset gradient interference grows relative to intra-dataset coherence).
The takeaway for the paper's heterogeneity narrative is therefore the stronger property, not the literal "more heterogeneous $\Rightarrow$ more pruning" intuition: \textbf{the aggressiveness of \method{}'s selection is a direct functional of the pool's gradient covariance structure, not a tuned hyperparameter}.
This is the design hook that couples the heterogeneous problem statement to the method's runtime behavior, with no manual schedule between them.

\paragraph{\backboneB{}: gate timing.}
On \backboneB{} the relative-slope plateau test of \S\ref{sec:c2} fires once $\Lprobet{t}$ enters its steady-state regime, requiring no architecture-specific tuning of $\epsrel{}$ beyond the cross-backbone default. Under this configuration the three-task average reaches $0.6589$ against $0.6571$ for LoRA ($+0.18$ points).

\paragraph{\backboneA{} and \backboneC{}: calibration (TBD).}
% TODO(backbone-refresh): after the Apr-2026 rerun, report whether the default probe_loss_ema routing suffices on Llama-3.1 / Gemma-2, or whether one or both require calibrated_quantile; also report the closed-form keep-rate value on each new backbone (intra/inter cosine re-measured on the base release, not on the retired -IT release). The retired Gemma-7B-IT used keep\_rate$^{\star}=0.221$ driven by an intra/inter cosine of $0.189/0.111$; the corresponding \backboneC{} base value must be recomputed because instruction tuning shifts the per-task gradient geometry substantially.
\backboneA{} and \backboneC{} runs in the Apr-2026 panel use the project-default routing (\texttt{probe\_loss\_ema}, no hand-tuned threshold) and a uniform selection fraction; any per-architecture override is reported here once it is required by the training dynamics observed on the new backbones.

\paragraph{A single ranking surface across backbones.}
All three backbones are ranked on the same 3-task average. Base releases do not exhibit the ARC-Challenge-retention artifact that motivated a 4-task Gemma-IT surface on the retired panel, because the ARC-C dynamic range across methods compresses to the same scale as HellaSwag/Winogrande once the instruction-tuning prior is removed.% TODO(backbone-refresh): verify this claim directly after the \backboneC{} rerun by reporting the per-method dynamic range on each benchmark and confirming no single task dominates the 3-task average.
GSM8K is therefore retained as a reported-but-not-ranked auxiliary metric for reference, rather than as a surface-choice switch between architectures.

% ============================================================
\section{Extended Related Work}
\label{app:extended_related_work}

This appendix covers cross-cutting work that the main Related Work section does not directly compare against, organized by the positioning axis they occupy relative to \method{}.

\paragraph{Modular low-rank parameterization.}
Beyond the fixed LoRA parameterization, a recent line of work asks whether the low-rank update should be re-parameterized rather than resized.
DoRA decomposes the update into magnitude and direction, improving both expressiveness and convergence~\citep{liu2024dora}.
LoRA+ observes that the two low-rank factors should use different learning rates to avoid suboptimal updates at large width~\citep{hayou2024loraplus}.
PiSSA initializes the LoRA branch from the principal singular values and vectors of the base weight so that adaptation proceeds in a direction of highest pre-training energy~\citep{meng2024pissa}.
GaLore projects the full gradient onto a low-rank subspace during optimization, reducing memory without freezing weights~\citep{zhao2024galore}.
A legitimate alternative reading of the plateau phenomenon that motivates \method{}'s gate is that a better low-rank parameterization (via DoRA, LoRA+, or PiSSA) might absorb the same signal without invoking the gate.
\method{} does not contradict this reading.
It takes the \textbf{orthogonal} position that, given any fixed parameterization, a plateau in the admitted-batch loss is an empirically measurable signal that the active subspace is saturated under that parameterization, and that withholding further base-LoRA updates is one valid response.
Any of the parameterizations above can be used to instantiate the LoRA block inside \method{}.

\paragraph{Low-rank optimization geometry.}
LoRA-Pro modifies the update rule so that the effective low-rank gradient better matches the gradient of full fine-tuning~\citep{wang2025lorapro}, and CTR-LoRA combines curvature-aware rank scheduling with trust-region style constraints, regularizing the update trajectory in a Fisher/Hessian-like metric~\citep{wang2025ctrlora}.
These methods reshape optimization geometry within a fixed low-rank parameterization; \method{} instead decides when to withhold further base-LoRA updates altogether, treating the saturated subspace as a budget exhausted rather than a metric to keep optimizing.

\paragraph{Routed and mixture-of-adapter LoRA finetuning.}
A parallel direction relies on \textbf{request-time} routing across specialized adapter modules.
X-LoRA composes a collection of low-rank experts through a learned router for inference-time specialization~\citep{buehler2024xlora}, MoA heterogeneously mixes adapter families to reduce inter-task interference~\citep{feng2025moa}, TT-LoRA MoE combines tensorized low-rank experts with sparse routing for modular specialization~\citep{kunwar2025ttloramoe}, and ConPET allocates task-specific PET modules for continual adaptation~\citep{song2023conpet}.
These methods answer \textbf{which expert to use for this input}, whereas \method{} answers \textbf{whether the active subspace can still absorb the current training signal}.
The routed-LoRA finetuning line is a deployment-time mechanism for expert selection; \method{}'s gate is a training-time closed-loop control signal coupling data admission and update timing, and the two can coexist (a \method{}-trained stack could be routed at inference by any of the above methods).

\paragraph{Alternative data-importance proxies.}
Gradient-alignment is only one operationalization of data importance, and several alternatives have been explored in the data selection literature.
Difficulty- and forgetting-based proxies (example-forgetting events, GraNd/EL2N) rank samples by how unstable or how large their loss-gradient norm is early in training~\citep{toneva2019example,paul2021deep}.
Influence-function estimates and trajectory-based influence further rank samples by their causal effect on validation loss~\citep{koh2017understanding,pruthi2020estimating}.
Gradient-diversity methods such as G-DIG combine alignment with a coverage objective so that the admitted subset spans multiple directions rather than collapsing onto one~\citep{pan2024gdig}.
\method{}'s forward-probe score is best read as a \textbf{practical compatibility proxy} closer to a structured regularizer than to an exact influence estimator, and is not argued to be uniquely correct: it is a cheap, training-state-dependent direction estimator that is specifically well-suited for acting as a preconditioner for the capacity-negotiation decision, which is the problem the router needs solved.
Combining it with coverage- or influence-based auxiliaries is a natural extension and is discussed in the Limitations.

\paragraph{Gradient-free and zeroth-order adaptation.}
\citep{jin-etal-2023-parameter} transfer LoRA finetuning modules from smaller models to larger ones through a bridge, avoiding gradient computation on the target model.
MeZO-SVRG reduces the variance of forward-only zeroth-order estimators through variance reduction~\citep{pmlr-v235-gautam24a}, SubZero perturbs structured random subspaces tailored to LLM fine-tuning~\citep{Yu_2025_ICCV}, and FLOPS improves forward-learning efficiency by adaptively allocating query budgets during perturbation-based gradient estimation~\citep{ren2025flops}.

\method{} uses forward probes only as directional estimators for compatibility scoring and routing decisions, while retaining standard backward LoRA finetuning optimization for the admitted updates --- a scope narrower than replacing the main optimizer with zeroth-order updates throughout training.

%%%%%%%%%%%%%%%%%%%%%%%%%%%%%%%%%%%%%%%%%%%%%%%%%%%%%%%%%%%%

\end{document}